\documentclass[journal]{IEEEtran}
\ifCLASSINFOpdf
\else
\fi
\usepackage{array}
\newcommand{\PreserveBackslash}[1]{\let\temp=\\#1\let\\=\temp}
\newcolumntype{C}[1]{>{\PreserveBackslash\centering}p{#1}}
\newcolumntype{R}[1]{>{\PreserveBackslash\raggedleft}p{#1}}
\newcolumntype{L}[1]{>{\PreserveBackslash\raggedright}p{#1}}
\usepackage{multirow}
\usepackage{amsfonts}
\usepackage{subfigure}
\usepackage{booktabs} %toprule, midrule, bottomrule
\usepackage{graphicx}
\usepackage{amsmath}
\usepackage{comment}
\usepackage{caption}
\usepackage{threeparttable}

\begin{document}
%
% paper title
% Titles are generally capitalized except for words such as a, an, and, as,
% at, but, by, for, in, nor, of, on, or, the, to and up, which are usually
% not capitalized unless they are the first or last word of the title.
% Linebreaks \\ can be used within to get better formatting as desired.
% Do not put math or special symbols in the title.
\title{Targeted Physical-World Attention Attack on Deep Learning Models in Road Sign Recognition}
%
%
% author names and IEEE memberships
% note positions of commas and nonbreaking spaces ( ~ ) LaTeX will not break
% a structure at a ~ so this keeps an author's name from being broken across
% two lines.
% use \thanks{} to gain access to the first footnote area
% a separate \thanks must be used for each paragraph as LaTeX2e's \thanks
% was not built to handle multiple paragraphs
%

\author{Xinghao~Yang,
        Weifeng~Liu,~\IEEEmembership{Senior Member,~IEEE,}
        Shengli~Zhang,~\IEEEmembership{Senior Member,~IEEE,}
        Wei~Liu,~\IEEEmembership{Senior Member,~IEEE,}
       Dacheng~Tao,~\IEEEmembership{Fellow,~IEEE}
\thanks{Xinghao Yang and Wei Liu are with the Advanced Analytics Institute, Faculty of Engineering and Information Technology, University of Technology Sydney, Australia. E-mail: Xinghao.Yang@student.uts.edu.au, Wei.Liu@uts.edu.au.}
\thanks{Weifeng Liu is with the School of Information and Control Engineering, China University of Petroleum (East China), Qingdao 266580, China. E-mail: liuwf@upc.edu.cn.}
\thanks{Shengli Zhang is with the College of Information Engineering, Shenzhen University, Shenzhen 518060, China. E-mail: slzhang.szu@gmail.com.}
\thanks{Dacheng Tao is with the UBTECH Sydney Artificial Intelligence Centre, University of Sydney, Darlington, NSW 2008, Australia, and also with School of Information Technologies, Faculty of Engineering and Information Technologies, University of Sydney, Darlington, NSW 2008, Australia. E-mail: dacheng.tao@sydney.edu.au.}
\thanks{Manuscript received April 19, 2005; revised August 26, 2015.}
\thanks{(Corresponding Author: Shengli Zhang and Wei Liu)}
}
% note the % following the last \IEEEmembership and also \thanks -
% these prevent an unwanted space from occurring between the last author name
% and the end of the author line. i.e., if you had this:
%
% \author{....lastname \thanks{...} \thanks{...} }
%                     ^------------^------------^----Do not want these spaces!
%
% a space would be appended to the last name and could cause every name on that
% line to be shifted left slightly. This is one of those "LaTeX things". For
% instance, "\textbf{A} \textbf{B}" will typeset as "A B" not "AB". To get
% "AB" then you have to do: "\textbf{A}\textbf{B}"
% \thanks is no different in this regard, so shield the last } of each \thanks
% that ends a line with a % and do not let a space in before the next \thanks.
% Spaces after \IEEEmembership other than the last one are OK (and needed) as
% you are supposed to have spaces between the names. For what it is worth,
% this is a minor point as most people would not even notice if the said evil
% space somehow managed to creep in.

% The paper headers
\markboth{Targeted Attention Attack on Deep Learning Models in Road Sign Recognition}%
{Shell \MakeLowercase{\textit{et al.}}: Bare Demo of IEEEtran.cls for IEEE Journals}
% The only time the second header will appear is for the odd numbered pages
% after the title page when using the twoside option.
%
% *** Note that you probably will NOT want to include the author's ***
% *** name in the headers of peer review papers.                   ***
% You can use \ifCLASSOPTIONpeerreview for conditional compilation here if
% you desire.

% If you want to put a publisher's ID mark on the page you can do it like
% this:
%\IEEEpubid{0000--0000/00\$00.00~\copyright~2015 IEEE}
% Remember, if you use this you must call \IEEEpubidadjcol in the second
% column for its text to clear the IEEEpubid mark.

% use for special paper notices
%\IEEEspecialpapernotice{(Invited Paper)}

% make the title area
\maketitle

% As a general rule, do not put math, special symbols or citations
% in the abstract or keywords.
\begin{abstract}
Real world traffic sign recognition is an important step towards building autonomous vehicles, most of which highly dependent on Deep Neural Networks (DNNs). Recent studies demonstrated that DNNs are surprisingly susceptible to adversarial examples. Many attack methods have been proposed to understand and generate adversarial examples, such as gradient based attack, score based attack, decision based attack, and transfer based attacks. However, most of these algorithms are ineffective in real-world road sign attack, because (1) iteratively learning perturbations for each frame is not realistic for a fast moving car and (2) most optimization algorithms traverse all pixels equally without considering their diverse contribution. To alleviate these problems, this paper proposes the targeted attention attack (TAA) method for real world road sign attack. Specifically, we have made the following contributions: (1) we leverage the soft attention map to highlight those important pixels and skip those zero-contributed areas – this also helps to generate natural perturbations, (2) we design an efficient universal attack that optimizes a single perturbation/noise based on a set of training images under the guidance of the pre-trained attention map, (3) we design a simple objective function that can be easily optimized, (4) we evaluate the effectiveness of TAA on real world data sets. Experimental results validate that the TAA method improves the attack successful rate (nearly 10\%) and reduces the perturbation loss (about a quarter) compared with the popular $\text{RP}_\text{2}$ method. Additionally, our TAA also provides good properties, e.g., transferability and generalization capability. We provide code and data to ensure the reproducibility: https://github.com/AdvAttack/RoadSignAttack.
\end{abstract}

% Note that keywords are not normally used for peerreview papers.

% For peer review papers, you can put extra information on the cover
% page as needed:
% \ifCLASSOPTIONpeerreview
% \begin{center} \bfseries EDICS Category: 3-BBND \end{center}
% \fi
%
% For peerreview papers, this IEEEtran command inserts a page break and
% creates the second title. It will be ignored for other modes.
\IEEEpeerreviewmaketitle

\section{Introduction}
% The very first letter is a 2 line initial drop letter followed
% by the rest of the first word in caps.
%
% form to use if the first word consists of a single letter:
% \IEEEPARstart{A}{demo} file is ....
%
% form to use if you need the single drop letter followed by
% normal text (unknown if ever used by the IEEE):
% \IEEEPARstart{A}{}demo file is ....
%
% Some journals put the first two words in caps:
% \IEEEPARstart{T}{his demo} file is ....
%
% Here we have the typical use of a "T" for an initial drop letter
% and "HIS" in caps to complete the first word.
\IEEEPARstart{I}{nternet} of Things (IoT) applications in smart traffic control and smart cities are highly dependent on intelligent autonomous vehicles \cite{9098906}\cite{6740844}\cite{6702523}. The traffic sign recognition is the core function of such IoT driver assistance systems \cite{8306879}\cite{8957685}\cite{nguyen2016trafficwatch}. Deep Neural Networks (DNNs) got the best performances in the 2011 traffic sign recognition competition \cite{stallkamp2011german}, and are likely to enable the future self-driving cars \cite{Kim_2019_CVPR}. However, recent studies have shown that DNNs tend to deliver a wrong prediction with high confidence if the input data are slightly perturbed \cite{szegedy2013intriguing}\cite{goodfellow2014explaining}\cite{braytee2016cost}\cite{Zeng_2019_CVPR}\cite{liu2011quadratic}. These perturbations are often imperceptible to human vision system, but their effects on DNNs are catastrophic \cite{akhtar2018threat}.

\begin{figure}[t]
\begin{center}
\includegraphics[height=1.85cm]{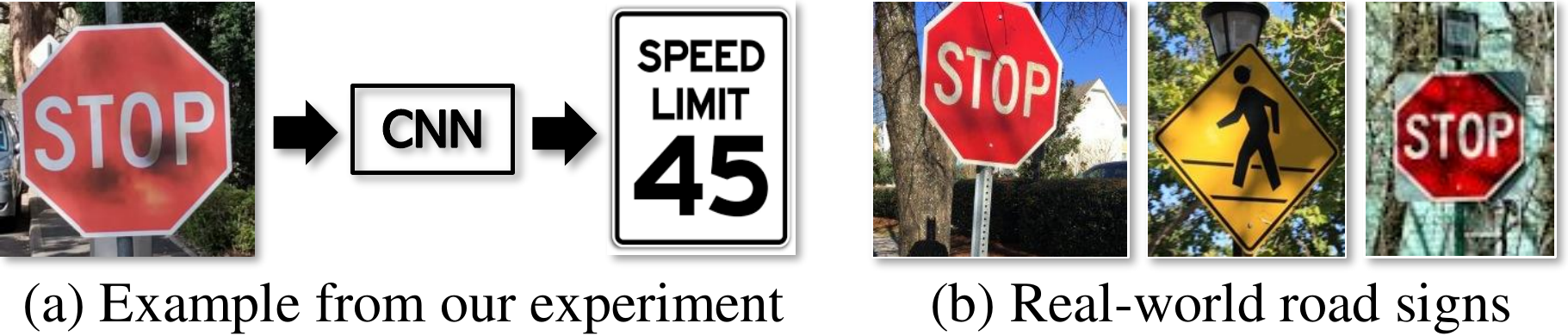}
\end{center}
\setlength{\abovecaptionskip}{0pt}
\caption{(a) In our experiment, the ``Stop" sign is perturbed by grayscale noises, then a well-trained CNN classifier gives a wrong prediction. (b) Three real-world road signs with tree shadows. Noises in (b) are similar to those in (a).}
\label{fig-Introductioin-TAA-Stop-Speedlimit45}
\end{figure}

This phenomenon poses great challenges to DNNs security implementation, especially in real world traffic sign recognition \cite{chen2015deepdriving}. For example, Fig.~\ref{fig-Introductioin-TAA-Stop-Speedlimit45} shows a ``Stop" sign which is attacked by grayscale perturbations. Such perturbations are naturally regarded as some graffiti or tree shadows by human driver, but they successfully mislead the well-trained Convolutional Neural Network (CNN) classifier to ``SpeedLimit45". This may cause irreparable loss of life and property, because the car will keep on moving at a limited speed instead of fully braking. Therefore, it is important to study the possible attack methods and make sure the self-driving car is capable of immunizing these attacks before launching.

Currently, a number of adversarial attack methods have been proposed, such as gradient based attack, score based attack, decision based attack and transformation based attacks (Section \ref{adversarial attack methods}). However, most of these attack methods are not strong enough in real world road sign attack due to the following two reasons. (1) These methods are designed for \textit{single} image attack using iterative optimization strategy. Imagine that a self-driving car is passing by a ``Stop" sign, and the camera captures a series of consecutive images in a short time. In this case, iteratively calculating the minimal perturbations for each frame can be infeasible. (2) In their optimization, all pixels must be processed either at once or one by one. They ignore the fact that pixels in different locations have different attack potentials \cite{su2019one,narodytska2016simple}.

To tackle the real world scenario, Eykholt \emph{et al.} \cite{eykholt2018robust} designed the robust physical perturbations ($\text{RP}_\text{2}$) method, which learns a single perturbation based on \textit{a set of} road sign images. In their model, the $\text{RP}_\text{2}$ first learns a sparse mask (with value 0 or 1) via L1-norm, and then adds perturbations to those 1-value-pixels using L2-norm. This perturbation achieves a robust attack performance when it is printed to a real traffic signpost. However, the L1 mask has some limitations: (1) it leads to a high perturbation cost since all attacked pixels are given the biggest weight, i.e., 1; (2) it needs manual steps to make the L1 mask looks like rectangle; (3) the perturbation in the rectangle shape is yet not natural enough.

In order to reduce the perturbation cost and achieve a robust attack performance, we propose the Targeted Attention Attack (TAA) method, which utilizes a soft attention map (with value between 0 and 1) to distinguish pixels weights. Specifically, TAA contains two main steps: the first step learns a soft attention map, which highlights those important pixels and suppresses trivial pixels in classification; the second step reduces the adversarial noises in terms of the $L_2$-norm according to the pre-learned attention map. We summarize the main contribution of this work as below:
\begin{itemize}
\item We propose an attacking strategy based on soft attention maps, which is associated with low cost and natural output - more rational to an attacker.
\item We design the Targeted Attention Attack (TAA) for real world attack. For efficiency, TAA constructs a single adversarial perturbation to mislead all test road signs captured under different distance and view-angles.
\item We formulate an objective function for the final attacks that make use of the soft attention maps and it can be easily optimized by existing optimizers.
\item We evaluate the feasibility of TAA on both existing datasets (LISA and GTSRB) and physical world road signs. Extensive experimental results show that TAA achieves a high targeted attack rate with small perturbations. It also exhibits strong transferability and generalization capability.
\end{itemize}

\section{Related work}
\label{RelatedWorks}
In this section, we first review four kind of adversarial attack methods, i.e., gradient based attack, score based attack, decision based attack and transformation based attack, and then discuss the $\text{RP}_\text{2}$ model and the attention mechanism.

\subsection{Adversarial attack methods}
\label{adversarial attack methods}
Firstly, gradient based attack \cite{goodfellow2014explaining,kurakin2016adversarial,moosavi2016deepfool,Dong2018CVPR,9149635} seeks for the most sensitive perturbing direction for an input data according to the gradient of loss function. Goodfellow \emph{et al.} \cite{goodfellow2014explaining} proposed the well-known fast gradient sign method (FGSM), which determines the perturbation direction (increase or decrease) for each pixel by leveraging the gradient of the loss function. FGSM is designed for fast learning via a single gradient step, so it often fails to find the minimal perturbation \cite{carlini2017towards}. Kurakin \emph{et al.} \cite{kurakin2016adversarial} refined the FGSM by repeating the gradient step many times with a smaller step size in each iteration. This iterative FGSM (I-FGSM) misleads the classifier in a higher rate with relatively smaller perturbations. The DeepFool method \cite{moosavi2016deepfool} further reduces the perturbations strength by iteratively searching for the distance between a clean input to its closest classification hyperplane. However, the greedy optimization strategies in both I-FGSM and DeepFool are easily leading to a local optimum. Dong \emph{et al.} \cite{Dong2018CVPR} designed the momentum I-FGSM (MI-FGSM), which employs a velocity vector to memorize all the previous gradients during iterations to escape from poor local maximum. Recently, Xiang \emph{et al.} \cite{9149635} embedded the FGSM into the gray-box attack scheme, where the victim network structure is inaccessible but can be derived by the side-channel attack (SCA).

Secondly, score based attack \cite{narodytska2016simple,hayes2017machine,ilyas2018black} relies on the output scores (e.g., predicted probability) instead of the gradient information in constructing adversarial perturbations, without access of either model architecture and model weights. Narodytska \emph{et al.} \cite{narodytska2016simple} applied the confidence score to guide a greedy local-search method, which finds a few pixels (even single pixel) that are most helpful in generating perturbations. It adopts the ``top-$k$ misclassification" criteria, which means the search procedure will stop until it pushes the correct label out of the top-$k$ scores. Hayes and Danezis \cite{hayes2017machine} trained a attacker neural network to learn perturbations, which then used to attack another block-box network. They defined the loss function by combining the output confidence scores of both networks. Ilyas \emph{et al.} \cite{ilyas2018black} considered more realistic settings, e.g., the attacker only knows a limited number of prediction scores due to the query limitation. The authors employed the natural evolution strategy (NES) to estimate the gradient and adopted the projected gradient descent to generate adversarial examples. Based on \cite{ilyas2018black} , Zhao \emph{et al.} \cite{zhao2020towards} derived the Fisher information matrix (FIM) and incorporated FIM with the second-order natural gradient descent (NGD) to achieve high query-efficiency.

Thirdly, decision based attack \cite{brendel2018decisionbased,schott2018towards,chen2019boundary,9152788,Li_2020_CVPR} requires only the model classification decision (i.e., the top-1 class label) and frees the need of either model gradient or their output scores. The boundary attack \cite{brendel2018decisionbased} starts with an adversarial point and then reduces the noise by implementing the random walk along the decision boundary. This method adds minimal perturbations (in terms of L2-distance) compared with gradient based methods, but it needs much more iterations to converge. The pointwise attack \cite{schott2018towards} initializes the starting point with salt-pepper noise. Then it repeatedly resets each perturbed pixel to clean image while making sure the noisy image still adversarial. Chen and Jordan \cite{chen2019boundary} developed the Boundary Attack ++, which not only reduces the number of model queries but also able to switch between $L_2$ and $L_\infty$ distance by designing two clip operators. In \cite{9152788}, Chen \emph{et al.} employed the binary information of the decision boundary to estimate the gradient direction and presented the decision based HopSkipJumpAttack (HSJA). This method achieved competitive results by attacking popular defense mechanisms, while its query efficiency needs improvement. Li \emph{et al.} \cite{Li_2020_CVPR} pointed out that the large number of query iterations for boundary-based attack is due to the high dimensional input (e.g., image). Thereby, three subspace optimization methods (i.e., spatial subspace, frequency subspace and priciple component subspace) is explored in their Query-Efficient Boundary-based blackbox Attack (QEBA) for perturbation sampling.

Finally, transformation based attack \cite{xiao2018spatially,engstrom2019exploring,wang2020deceiving,chen2020explore} crafts adversarial images by shifting pixels' spatial location instead of directly modifying their value. For example, Xiao \emph{et al.} \cite{xiao2018spatially} proposed the spatially transformed adversarial (stAdv) method, which replaces the $L_p$-norm with local geometry distortion in measuring the magnitude of perturbations. The stAdv minimizes the transformation perturbation with L-BFGS optimizer \cite{liu1989limited}. Engstrom \emph{et al.} \cite{engstrom2019exploring} also found that simply rotating or translating a natural image is enough to fool a deep vision model, even though the solution may be not the global optimal. Wang \emph{et al.} \cite{wang2020deceiving} investigated the effect of image spatial transformation on the image-to-image (Im2Im) translation task, which is more sophisticated than pure classification problem.  They revealed that the geometrical image transformation (i.e., translation, rotation and scale) in the input domain can cause incorrect color map of Im2Im framework in the target domain. Different from the previous works that depend only on the spatial transformation, Chen \emph{et al.} \cite{chen2020explore} integrated linear spatial transformation (i.e., affine transformation) with color transformation and proposed a two-phase combination attack. These adversary models can be potentially applied to protect social users' interaction for influence learning \cite{LI2020101522,cai2020target}.

Based on attacker's knowledge, these methods can be divided into white-box attack, black-box attack and gray-box attack. Specifically, white-box attack assumes attackers know everything about the victim model (e.g., architecture, parameters, training method and data); black-box attack assumes the adversary only knows the output of the model (prediction label or probability) given an input; and gray-box attack means the scenario where the hacker knows part of information (e.g., the network structure) and the rest (e.g., parameters) is missing. Based on attacker's specificity, these methods fall into targeted attack where the model outputs a user specified label; or untargeted attack where the model is misled to any label other than the correct label. A summary is provided in Table \ref{tab-summary}.
\begin{table}[t]
\center
\caption{Summary of the properties for different attacking methods. The properties are \textbf{T}argeted attack, \textbf{U}ntargeted attack, \textbf{W}hite-box attack, \textbf{B}lack-box attack and \textbf{G}ray-box attack.}
\label{tab-summary}
\begin{tabular}{ l  p{0.63cm}<{\centering}  p{0.63cm}<{\centering}  p{0.63cm}<{\centering}  p{0.63cm}<{\centering} p{0.63cm}<{\centering} }
\toprule
\multirow{2}*{\textbf{Attacking Methods}}  & \multicolumn{5}{c}{\textbf{Properties}} \\
\cmidrule(lr){2-6}
~ & \textbf{T}  & \textbf{U} & \textbf{W} & \textbf{B} & \textbf{G} \\
\midrule
FGSM \cite{goodfellow2014explaining}                 & $\checkmark$ & $\checkmark$ & $\checkmark$ &               &    \\
I-FGSM \cite{kurakin2016adversarial}                 & $\checkmark$ & $\checkmark$ & $\checkmark$ &               &    \\
DeepFool \cite{moosavi2016deepfool}                  &              & $\checkmark$ & $\checkmark$ &               &    \\
MI-FGSM \cite{Dong2018CVPR}                          & $\checkmark$ & $\checkmark$ & $\checkmark$ & $\checkmark$  &    \\
Xiang \emph{et al.} \cite{9149635}     & $\checkmark$ &              &              &               &  $\checkmark$  \\
Narodytska \emph{et al.} \cite{narodytska2016simple} & \multicolumn{2}{c}{Top-k misclass}   & $\checkmark$ &               &    \\
Hayes and Danezis \cite{hayes2017machine}            & $\checkmark$ & $\checkmark$ &              & $\checkmark$  &    \\
Ilyas \emph{et al.} \cite{ilyas2018black}            & $\checkmark$ &              &              & $\checkmark$  &    \\
Zhao \emph{et al.} \cite{zhao2020towards} &           & $\checkmark$ & $\checkmark$ &               &    \\
Boundary Attack \cite{brendel2018decisionbased}      & $\checkmark$ & $\checkmark$ &              & $\checkmark$  &    \\
Pointwise Attack \cite{schott2018towards}            &              & $\checkmark$ &              & $\checkmark$  &    \\
Boundary Attack ++ \cite{chen2019boundary}           & $\checkmark$ & $\checkmark$ &              & $\checkmark$  &    \\
HSJA \cite{9152788}                    & $\checkmark$ & $\checkmark$ &              & $\checkmark$  &    \\
QEBA \cite{Li_2020_CVPR}               & $\checkmark$ &              &              & $\checkmark$  &    \\
stAdv \cite{xiao2018spatially}                       & $\checkmark$ &              & $\checkmark$ &               &    \\
Engstrom \emph{et al.} \cite{engstrom2019exploring}  &              & $\checkmark$ &              & $\checkmark$  &    \\
Wang \emph{et al.} \cite{wang2020deceiving}   &       & $\checkmark$ &              & $\checkmark$  &    \\
Chen \emph{et al.} \cite{chen2020explore}   & $\checkmark$ &         & $\checkmark$ &               &    \\
\bottomrule
%\multicolumn{6}{l}{$^{\mathrm{1}}$\textbf{T}: Targeted, \textbf{U}: Untargeted, \textbf{W}: White-box, \textbf{B}: Black-box, \textbf{G}: Gray-box.}
\end{tabular}
\end{table}

\subsection{$\text{RP}_\text{2}$ and attention mechanism}
Suppose we are given a clean example $x$ with the ground truth label $y$ and a classifier $f_\theta\left(\cdot\right)$ with well-trained parameter $\theta$, so we have $f_\theta\left(x\right)=y$. The goal of an attacker is to generate the adversarial example $x'$ who looks similar to $x$ but misleads the classifier to any wrong label $f_\theta\left(x'\right)\neq y$ (untargeted attack) or a pre-specified label $f_\theta\left(x\right)=y^*\neq y$ (targeted attack).

The $\text{RP}_\text{2}$ \cite{eykholt2018robust} is designed for generating robust perturbations for physical world road sign. Specifically, it learns a perturbation $\delta$ for an input ``Stop" sign $o$ and prints the perturbed sign on paper, with the hope that the attacked real object could be misclassified as ``SpeedLimit45". To ensure the attack is robust - the physical sign could consistently mislead the classifier under diverse distances and view angles, the perturbation $\delta$ is learned from a set of stop signs $X^V$. Formally, the optimization problem is:
\begin{eqnarray}
\label{equ-train}
\begin{aligned}
\mathop{\arg\min}_{\delta} & \lambda||M_x\cdot\delta||_p + NPS \\
& + \mathbb{E}_{x_i\sim X^V}J\left( f_\theta\left( x_i + R_i\left( M_x\cdot\delta \right) \right), y^* \right)
\end{aligned}
\end{eqnarray}
where $NPS$ is the non-printability score to evaluate the colour printer's fabrication error \cite{sharif2016accessorize}. $\lambda$ is a trade-off hyperparameter to balance the $p$-norm of perturbation and the loss function $J\left(\cdot,\cdot\right)$. The $R_i$ is a rotation alignment function, which ensures the perturbation could rotate if the object $x_i$ is rotated. The $\text{RP}_\text{2}$ performs the attention attack via the perturbation mask $M_x$, which assigns one to sensitive pixels and zero to others. Thus, only the sensitive area will be processed during the optimization. However, the mask $M_x$ is not a purely learning result, it needs some expert knowledge and manual intervention. Specifically, $M_x$ has two shapes: the first one is a hardcoded ``octagon" to match the shape of ``Stop" sign; the second one is a sparse matrix learned by $L_1$-norm but needs human modification to imitate a sticker attack. Reference Figure \ref{fig-processL1mask} for more details.

Wang \emph{et al.} \cite{wang2017residual} presented the Residual Attention Network (RAN), which can be easily extended to very deep (i.e., hundreds of layers) by directly stacking attention modules. In RAN, each attention module is composed of two branches: a trunk branch $T$ runs the feature processing and a soft mask branch $M$ acts as a feature selector. The mask branch was designed in a bottom-up top-down structure to mimic human cortex path \cite{mnih2014recurrent}.

Specifically, given an input image $x$, the output of trunk branch $T\left( x \right)$ and mask branch $M\left( x \right)$ have the same size. Instead of directly performing the element-wise product as:
\begin{eqnarray}
\label{equ-HighwayNetwork}
H\left( x \right) = M\left( x \right) \cdot T\left( x \right)
\end{eqnarray}
the RAN employs the attention residual learning by:
\begin{eqnarray}
\label{equ-ResidualLearning}
H\left( x \right) = \left( 1 + M\left( x \right) \right) \cdot T\left( x \right)
\end{eqnarray}
to avoid the following two problems: (1) repeatedly dot product with $M\left( x \right)$, whose value range between $[0,1]$, will continuously degrade the feature value in deep layers; (2) the identical mapping cushions the effect that the soft mask may break some good features.

\section{Targeted Attention Attack}
\label{Algorithm}
In this section, our Targeted Attention Attack (TAA) algorithm is formulated. TAA is designed for attacking real-world road signs by simulating human perception process. Before elaborating on the TAA, we first discuss the cornerstone challenges and difficulties that should be addressed in our model.

\subsection{Challenges from real-world conditions}
\label{Subsection3.1}
A major aim of attacking real-world road sign images is to mislead autonomous vehicles. Recall the real scene that a self-driving car passes by a ``Stop" sign and collects a number of digital images by a moving camera. It is nearly infeasible for an adversarial to iteratively optimize the perturbations for every image before they are fed into a classifier. In this case, we aim to learn a ``universal" perturbation for all collected images instead of implementing image-specific attack. However, the same ``Stop" sign will have great differences when it is photographed from different distances and view angles. Some special background and light even lead to the road sign being hardly visible. An ideal attack method should be effective under different environmental conditions.

\subsection{Target Attention Attack Work-flow}
\label{Subsection3.2}
\begin{figure*}[t]
\begin{center}
\includegraphics[height=3.6cm]{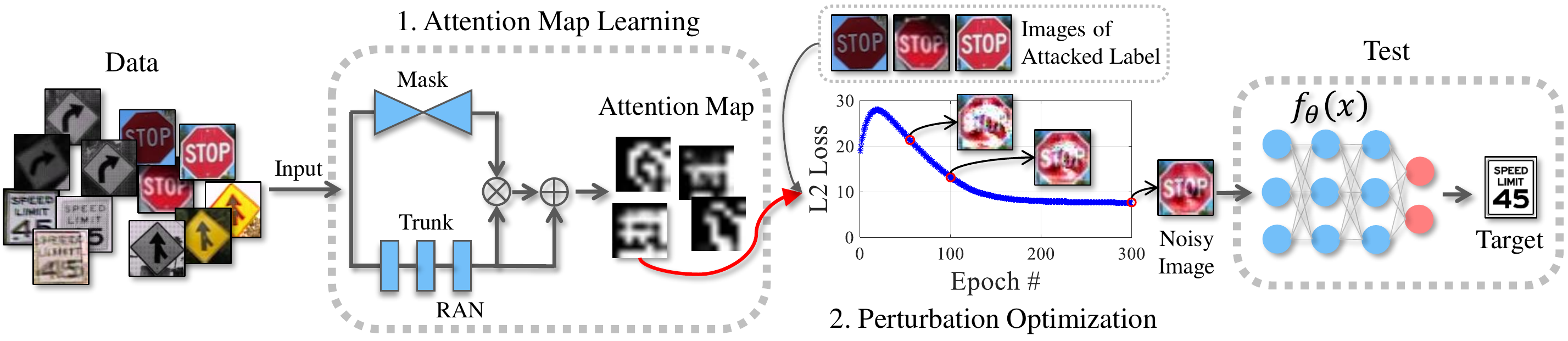}
\end{center}
\setlength{\abovecaptionskip}{0pt}
\caption{The TAA framework mainly contains two steps: soft attention map learning and perturbation optimization. The soft attention map is learned via a 92-layer RAN. The perturbation is optimized using the \emph{target} attention map and a set of training images. The red arrow means if our target is ``SpeedLimit45", then TAA takes the soft attention map from ``SpeedLimit45" instead of ``Stop".}
\label{fig-framework}
\end{figure*}

Fig.~\ref{fig-framework} illustrates the framework of TAA. In the left part, TAA takes the advantage of RAN to extract the soft attention map, which yields a smaller perturbation and higher fooling rate than the $L_1$-mask used in $\text{RP}_\text{2}$ \cite{eykholt2018robust}. In the right part, TAA uses a set of training images to optimize the perturbation, instead of basing on a single image \cite{goodfellow2014explaining,brendel2018decisionbased}, to fit the real world case. In our modelling, we have $N$ road sign images $X=\left\{ x_i\in\Re^{m \times n} \right\}_{i=1}^N$, belonging to $L$ classes $X=\left\{ X_l \right\}_{l=1}^L$.

\subsubsection{Soft attention map learning} In RAN \cite{wang2017residual}, each attention module is composed of a trunk branch and a soft mask branch. In practice, RAN stacks multiple attention modules to extract image feature for classification. Increasing attention modules of RAN could deliver a constant performance improvement owing to the residual structure \cite{wang2017residual}. In this work, we use the 92-layer RAN (attention-92), which contains 3 attention stages, stacking $\left\{ 1,2,3 \right\}$ attention modules, separately. So we totally have 6 attention modules. Take the first attention module as an example: given an input $x_i^1$\footnote{Here we denote the input of the first attention module as $x_i^1$ instead of $x_i$ because the raw image $x_i$ should first pass through a residual unit.}, its output is:
\begin{eqnarray}
\label{equ-ResidualModule1}
H_{1,c}\left( x_i^1 \right) = \left( 1 + M_{1,c}\left( x_i^1 \right) \right) \cdot T_{1,c}\left( x_i^1 \right)
\end{eqnarray}
where subscript $\left\{ 1,c \right\}$ denotes the $1^{st}$ attention module and the $c^{th}$ channel with $c\in [1,2,\ldots,C_1]$. The output of the first attention module then fed as the input of the second attention module after an residual unit \cite{he2016identity}. The whole RAN framework can be trained end-to-end.

After training, different attention modules carry out diverse attention performance. For example, low-level attention modules just reduces background feature, while high-level attention modules focus more on the part features that really matters for classification.  To achieve a precise attack, we select the attention map from the last attention module, i.e., $H_{j,c}\left( x_i^j \right)$, where $j=6$ in this work. We empirically found that $H_{j,c}\left( x_i^j \right)$ has a higher discrimination than $M_{j,c}\left( x_i^j \right)$, so the $H_{j,c}\left( x_i^j \right)$ is used as the soft attention map. By setting the channel numbers of the last attention stage, i.e., $C_4=C_5=C_6=1$, we can obtain $N$ attention maps $\left\{H_j\left( x_i^j \right)\right\}_{i=1}^N$, corresponding to $N$ training samples.

Before perturbation optimization, we should select only one attention map for each class and resize them into $m \times n$. Considering the real-world case that uses one perturbation to attack one class, the attention map should be the same for each class. In this paper, we select the one that is the closest to the average map in terms of the Euclidean distance
\begin{eqnarray}
\label{equ-findclosestmap}
\mathop{\arg\min}_{H_j\left( x_i^j \right),x_i \in X_l} \|Ave\left( H_j\left( X_l^j \right) \right) - H_j\left( x_i^j \right)\|_2
\end{eqnarray}
where $X_l$ is the subset of $X$, containing all the $x_i$ belong to the $l^{th}$ class. $Ave\left( \cdot \right)$ returns the average map. Let $H\left( X_l^{opt} \right)$ be the optimal attention map of the $l^{th}$ class. It is worth mentioning that there may be other potential methods for attention map selection. Finally, these $L$ attention maps are resized to $m \times n$ via bilinear interpolation
\begin{eqnarray}
\label{equ-bilinear}
A_l = Bilinear\left( H\left( X_l^{opt} \right) \right), l=1 \ldots L
\end{eqnarray}
where $A_l\in\Re^{m \times n}$ represents the soft attention map of class $l$. In $A_l$, different area have different attention weight, ranging from zero to one, to distinguish their contribution in classification. During optimization, an ideal attacker should pay more attention to those critical pixels and jump over those trivial ones.

\subsubsection{Perturbation optimization} In this part, we first derive the road sign attack without using attention mechanism, and then formulate our TAA model. In real-world cases, an targeted attacker aims to learn an universal perturbation from a set of training images, and hopes this perturbation would be effective for all the newly collected images, i.e., test images. Formally, let $X_l^{train}$ and $X_l^{test}$ be the training set and testing set of $l^{th}$ class, respectively. In training phase, the perturbation $\delta$ should be reduced to be as small as possible, but can still mislead the well trained classifier $f_\theta\left( \cdot \right)$ to a pre-specified class $t$:
\begin{eqnarray}
\begin{aligned}
\mathop{\arg\min}_{\delta,x_i\in X_l^{train}} D\left( x_i, x_i+\delta \right) \\
s.t. \ f_\theta\left( x_i+\delta \right) = y^*
\end{aligned}
\end{eqnarray}
where $t=\mathop{\arg\max}\left( y^* \right)$ and $D$ is a distance function. This optimization problem can be solved by reshaping to a relaxed Lagrange function \cite{eykholt2018robust}:
\begin{eqnarray}
\label{equ-lagrange}
\mathop{\arg\min}_{\delta} \lambda||\delta||_p + \mathbb{E}_{x_i\sim X_l^{train}}J\left( f_\theta\left( x_i+ \delta \right), y^* \right)
\end{eqnarray}
where $J\left( \cdot,\cdot \right)$ is loss function and $L_p$-norm measures the perturbation loss. In this work, the $NPS$ term in Eq. (\ref{equ-train}) is not used because it does not make a significant performance improvement and yet computationally expensive as discussed in \cite{sharif2016accessorize}.

TAA performs attention based attack by implementing element-wise product between the attention map $A_t$ and perturbation $\delta$, i.e., $A_t\cdot\delta$. Thus the optimization problem becomes:
\begin{align}
\mathop{\arg\min}_{\delta} \ \lambda||A_t\cdot\delta||_p + \mathbb{E}_{x_i\sim X_l^{train}}J\left( f_\theta\left( x_i+ A_t\cdot\delta \right), y^* \right)
\end{align}
\begin{comment}
\begin{equation}
\label{equ-TAA}
\begin{split}
&\mathop{\arg\min}_{\delta} \ \lambda||A_t\cdot\delta||_p \\
& \qquad + \mathbb{E}_{x_i\sim X_l^{train}}J\left( f_\theta\left( x_i+ A_t\cdot\delta \right), y^* \right)
\end{split}
%\nonumber
\end{equation}
\end{comment}
Here $A_t$ is the attention map of the target class. The $\delta$ is initialized randomly and then minimized using the ADAM optimizer\footnote{ADAM parameters: $\beta_1=0.9$, $\beta_2=0.999$, $\epsilon=10^{-8}$.}.

\section{Experiments}
\label{Experiments}
In this section, we evaluate the attack performance of TAA. Section \ref{LISA dataset} introduces the two datasets and data preprocessing. Then the experimental settings and results are discussed in section \ref{Experimental settings} and section \ref{Experimental results}, respectively.

\subsection{Datasets and Preprocess}
\label{LISA dataset}
The LISA dataset \cite{mogelmose2012vision} consists of 47 classes of US road sign images including 7855 annotations on 6610 frames with varying distance and view angles. The raw images are collected by different cameras on moving cars, and their size rang from $640 \times 480$ to $1024\times 522$ pixels. Notably, the LISA dataset is not well balanced. For example, it contains 1821 ``Stop" signs but only 5 ``thruMergeLeft" signs. In our experiments, we select the road sign types that over 40 images, i.e., the top 26 classes. Finally, the selected 7505 annotations are resized to $32 \times 32$.

The GTSRB \cite{stallkamp2012man} is a German road sign dataset, which contains 43 classes with totally 51,840 images. In our experiments, we only use its ``Stop" and ``PedestrianCrossing" sign to test the transferability of the learned perturbation by $\text{RP}_\text{2}$ and our TAA. Precisely, 240 ``Stop" sign and 60 ``PedestrianCrossing" sign are selected from the test subset of GTSRB, which are also resized to $32 \times 32$.

\subsection{Experimental Settings}
\label{Experimental settings}
\textbf{Network Training.} The CNN model consists of three convolutional layers with each followed by a ``Relu" function. After flatten and dense processing, the final feature is classified by Softmax. To learn a single perturbation, the TAA should train the RAN and the CNN classifier separately. For each network, we randomly select $80\% (6004)$ samples for training and the rest $20\% (1501)$ images for testing. After training, the RAN and CNN classifier achieves $98\%$ and $94.1\%$ accuracy on test images, respectively.

\textbf{Evaluation Methods.} We use two methods to evaluate the attack performance, i.e., perturbation loss and attack success rate. The perturbation loss measures the amount of the perturbations in terms of its $L_2$-norm, i.e., $P_{loss} = ||\delta||_2$.
The attack success rate indicates the ability that how much an adversary can mislead the CNN classifier. Similar to our baseline $\text{RP}_\text{2}$, an successful targeted attack is defined as: an classifier can correctly predict an clean input $f_\theta \left( x_i \right)=y$, but it will be misled to an target class by a noisy image $f_\theta\left( x_i+\delta \right) = y^*$. Thus, the attack success rate ($ASR$) is
\begin{eqnarray}
\label{equ-attacksuccessrate}
ASR= \frac{\sum_{x_i\in X_l} \left\{ f_\theta \left( x_i \right)=y \wedge f_\theta\left( x_i+\delta \right) = y^* \right\}}{\sum_{x_i\in X_l} \left\{ f_\theta \left( x_i \right)=y \right\}}
\end{eqnarray}
Rationally, an attacker hopes to accomplish a high $ASR$ with a small $P_{loss}$.

\textbf{Comparison Methods.} We compare our TAA with the baseline $\text{RP}_\text{2}$ \cite{eykholt2018robust} and six single image attack methods, including both untargeted salt-pepper attack \cite{chan2005salt}, contrast reduction attack \cite{rauber2017foolbox}, Gaussian blur attack \cite{reinhard2010high} and targeted pointwise attack \cite{schott2018towards}, FGSM \cite{goodfellow2014explaining}, boundary attack \cite{brendel2018decisionbased}.
\begin{itemize}
\item Salt-pepper attack \cite{chan2005salt} increases the amount of salt and pepper noise until the single input is misclassified.
\item Contrast reduction attack \cite{rauber2017foolbox} reduces the contrast of the input using a linear search to find the smallest adversarial perturbation.
\item Gaussian blur attack \cite{reinhard2010high} blurs the inputs using a Gaussian filter with linearly increasing standard deviation.
\item Pointwise attack \cite{schott2018towards} starts with the salt and pepper noise and minimizes the adversarial perturbation using $L_0$ norm.
\item FGSM \cite{goodfellow2014explaining} computes the gradient of the loss function once and then seeks the minimum step size to craft adversarial samples.
\item Boundary attack \cite{brendel2018decisionbased} is the most popular decision based attack, which minimizes the $L_2$ norm of the perturbations to ensure the invisibility.
\item $\text{RP}_\text{2}$ \cite{eykholt2018robust} is an universal attack algorithm, which optimizes a common perturbation with the guidance of $L_1$ mask. Then the perturbation magnitude is reduced to a small $L_2$ loss.
\end{itemize}

The code for $\text{RP}_\text{2}$ attack is available from Github\footnote{\url{https://github.com/evtimovi/robust_physical_perturbations}}, and the parameters are set as default \cite{eykholt2018robust}. We implement the FGSM \cite{goodfellow2014explaining} based on the CleverHans v3.0.1\footnote{\url{https://github.com/tensorflow/cleverhans}}, and run the rest methods on Foolbox v1.9.0\footnote{\url{https://github.com/bethgelab/foolbox}}. All parameters are set as the recommended value.

\subsection{Experimental Results}
\label{Experimental results}
\subsubsection{Stop-SpeedLimit45}
\label{Stop-SpeedLimit45}
Similar to $\text{RP}_\text{2}$, we use perturbed ``Stop" sign to mislead a well-trained CNN classifier to ``SpeedLimit45". In this experiment, we randomly select $80\%(1457)$ ``Stop" signs for perturbation optimization. Then the single perturbation $\delta$ is added to all the remainder $20\%(364)$ test images to evaluate the attack performance. For fair comparison, we run those single image attack methods on all the training images to optimize $1457$ perturbations. Then we compute a single average perturbation to attack the testing images.

Table \ref{table:stop-speeedlimit45} summarizes the attack success rate and perturbation loss of the ``Stop-SpeedLimit45" attack. From Table \ref{table:stop-speeedlimit45} we know that the proposed TAA achieves the highest targeted attack rate on the 364 testing images. Although single image attack based methods optimize relatively small perturbation (except FGSM), they produce very low $ASR$ on the test set. This confirms our previous point that one-to-one attack strategy is powerless in real world road sign attack. Comparing with $\text{RP}_\text{2}$, our TAA not only improves the attack success rate, but also reduces the magnitude of perturbations.

Table \ref{table:stop-speeedlimit45} right part visualizes the attack results. The Original column displays the selected ``Stop" sign from training set. The Mask column shows the starting point (for point-wise and boundary attack) or corresponding mask images (for $\text{RP}_\text{2}$ and TAA). The final two columns are attack results that perturbed by either the average noises of all training images (Adv-all) or the image-specific noises (Adv-1). After averaging, most single image attack methods crafts nearly human imperceptible perturbations, and the universal attack strategy generates human visible noises. However, these noise cannot deceive human eyes in our real life, e.g., we may regard TAA noisy image as tree shadow or graffiti. Therefore, these noisy road signs will not pose a barrier to human driving, but may cause great degradation to autonomous vehicles.
\begin{table*}[t]
\begin{center}
\caption{``Stop-SpeedLimit45" attack results.}
\label{table:stop-speeedlimit45}
\begin{tabular}{L{3.3cm}m{1.3cm}<{\centering}m{1.3cm}<{\centering}m{1.3cm}<{\centering}m{1.3cm}<{\centering}m{1.3cm}<{\centering}m{1.3cm}<{\centering}}
\toprule\noalign{\smallskip}
Methods & $P_{loss}$ & $ASR$ & Original & Mask & Adv-all & Adv-1\\
\noalign{\smallskip}
\midrule
\noalign{\smallskip}
Salt-Pepper \cite{chan2005salt}                 & \textbf{1.83}      & 0.00\% &\includegraphics[height=0.8cm]{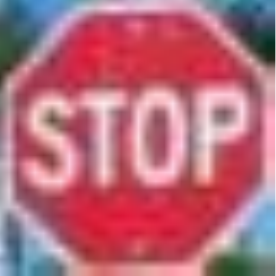}   & --- & \includegraphics[height=0.8cm]{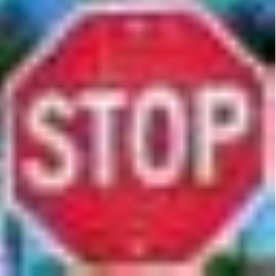} & \includegraphics[height=0.8cm]{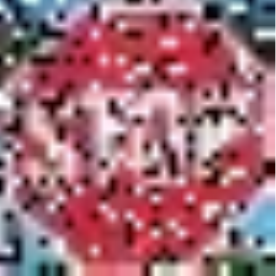} \\
Contrast reduction \cite{rauber2017foolbox}       & 5.90               & 0.00\% &\includegraphics[height=0.8cm]{fig-cleanstopsign.pdf}   & --- & \includegraphics[height=0.8cm]{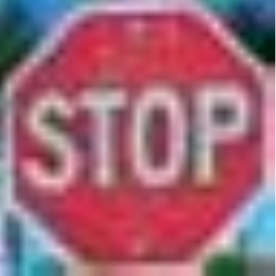} & \includegraphics[height=0.8cm]{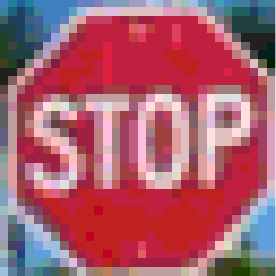} \\
Gaussian blur \cite{reinhard2010high}             & 2.93               & 0.00\% &\includegraphics[height=0.8cm]{fig-cleanstopsign.pdf}   & --- & \includegraphics[height=0.8cm]{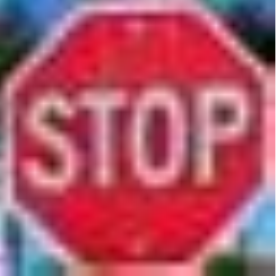} & \includegraphics[height=0.8cm]{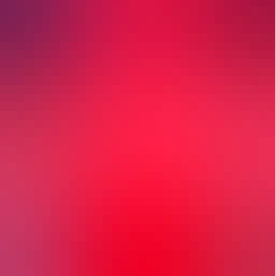} \\
FGSM \cite{goodfellow2014explaining}             & 14.20           & 47.50\%         &\includegraphics[height=0.8cm]{fig-cleanstopsign.pdf}   & --- & \includegraphics[height=0.8cm]{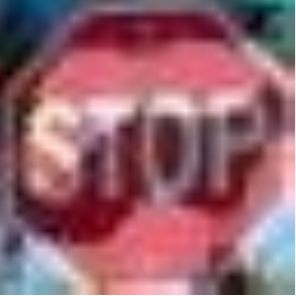} & \includegraphics[height=0.8cm]{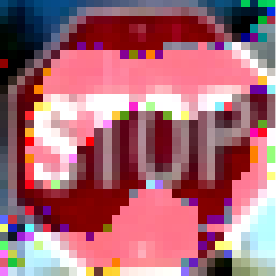} \\
Pointwise \cite{schott2018towards}              & 2.71               & 20.30\% &\includegraphics[height=0.8cm]{fig-cleanstopsign.pdf}   & \includegraphics[height=0.8cm]{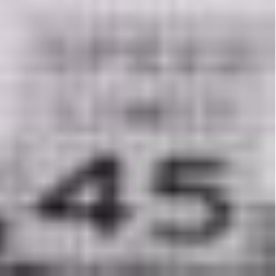} & \includegraphics[height=0.8cm]{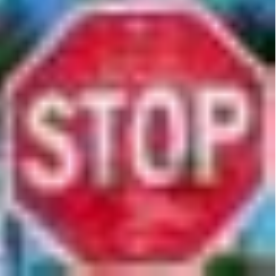} & \includegraphics[height=0.8cm]{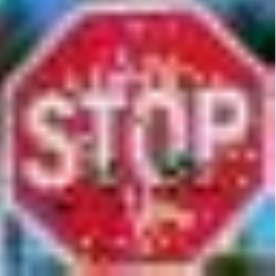} \\
Boundary attack \cite{brendel2018decisionbased}  & 1.92   & 23.90\%         &\includegraphics[height=0.8cm]{fig-cleanstopsign.pdf}   & \includegraphics[height=0.8cm]{fig-SpeedLimit45.pdf} & \includegraphics[height=0.8cm]{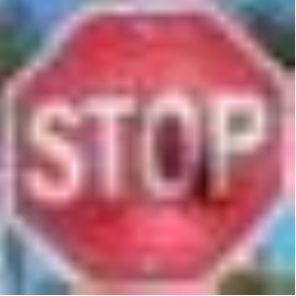} & \includegraphics[height=0.8cm]{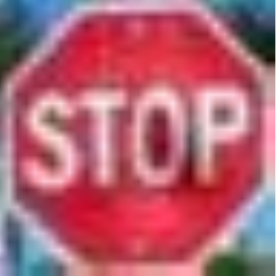} \\
$\text{RP}_\text{2}$ \cite{eykholt2018robust}                     & 10.81           & 91.80\%         &\includegraphics[height=0.8cm]{fig-cleanstopsign.pdf}   & \includegraphics[height=0.8cm]{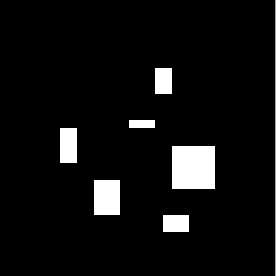} & \includegraphics[height=0.8cm]{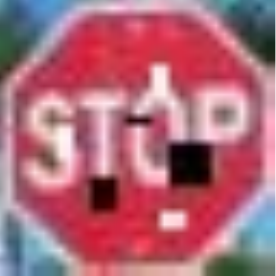} & --- \\
Our TAA                                          & 7.62            & \textbf{100\%}  &\includegraphics[height=0.8cm]{fig-cleanstopsign.pdf}   & \includegraphics[height=0.8cm]{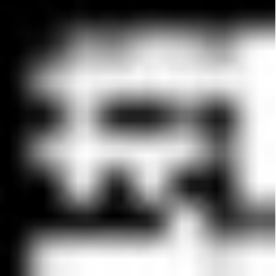} & \includegraphics[height=0.8cm]{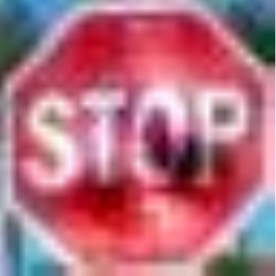} & --- \\
\bottomrule
\end{tabular}
\end{center}
\end{table*}

\subsubsection{PedestrianCrossing-Speedlimit65} In this part, we consider another dangerous attack - misclassifying a ``PedestrianCrossing" as a ``SpeedLimited65". In this case, the car will directly going through in a limited speed below 65 instead of slowing down to watch pedestrians. Thus this attack puts pedestrians in a very dangerous environment. Similarly, we randomly select $80\%(868)$ ``PedestrianCrossing" images to train both $\text{RP}_\text{2}$ and TAA, and repeatedly conduct six single image attack methods on every training image. Then their attack performance are evaluated on the other $20\%(217)$ images.

Table~\ref{table:pedestrian-speeedlimit65} reports the attack successful rate and perturbation $L_2$-loss. Similar to the ``Stop-SpeedLimit45" attack, Salt-Pepper adds smallest perturbations but our TAA achieves the highest attack accuracy. Besides, the $ASR$ column double confirms that single images based methods are not suitable for real world road sign attack. We will discard these algorithms in later parts. Comparing with $\text{RP}_\text{2}$, TAA gets a better performance again in terms of both attack success rate and perturbation magnitude. Table \ref{table:pedestrian-speeedlimit65} right part exhibits an example of ``PedestrianCrossing" sign and its noisy images. The original image is not very clear due to the tree shadow. Intriguingly, these shadow looks like our TAA perturbations to some extent. So we can explain the gradually changed white and black TAA perturbations more naturally, which imitates the change of real world shadow and sunlight. It can be perceived that $\text{RP}_\text{2}$ attack is very obviously due to the zero-one mask, while our TAA uses a soft attention mask so the perturbation appears more naturally. Furthermore, the soft attention mask should focus on less pixels with a deeper RAN structure in theory \cite{wang2017residual}.
\begin{table*}[t]
\begin{center}
\caption{``PedestrianCrossing-SpeedLimit65" attack results.}
\label{table:pedestrian-speeedlimit65}
\begin{tabular}{L{3.3cm}m{1.3cm}<{\centering}m{1.3cm}<{\centering}m{1.3cm}<{\centering}m{1.3cm}<{\centering}m{1.3cm}<{\centering}m{1.3cm}<{\centering}}
\toprule\noalign{\smallskip}
Methods & $P_{loss}$ & $ASR$ & Original & Mask & Adv-all & Adv-1 \\
\noalign{\smallskip}
\midrule
\noalign{\smallskip}
Salt-Pepper \cite{chan2005salt}                 & \textbf{1.44}      & 0.50\% &\includegraphics[height=0.8cm]{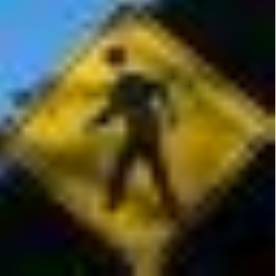}   & --- & \includegraphics[height=0.8cm]{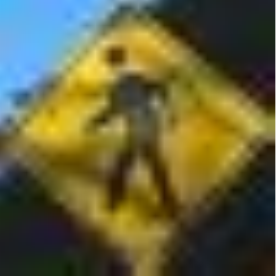} & \includegraphics[height=0.8cm]{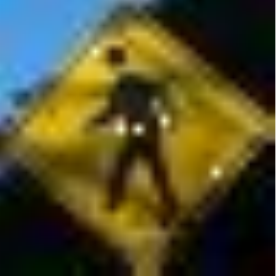}\\
Contrast reduction \cite{rauber2017foolbox}      & 4.56               & 2.30\% &\includegraphics[height=0.8cm]{fig-pedestrian-clean.pdf}   & --- & \includegraphics[height=0.8cm]{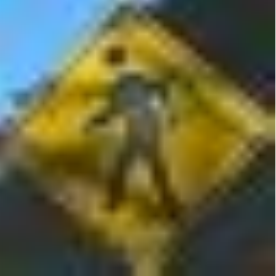} & \includegraphics[height=0.8cm]{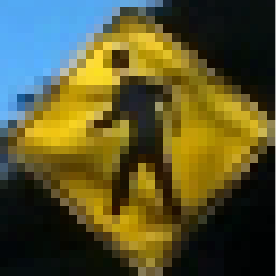} \\
Gaussian blur \cite{reinhard2010high}                       & 3.15               & 0.50\% &\includegraphics[height=0.8cm]{fig-pedestrian-clean.pdf}   & --- & \includegraphics[height=0.8cm]{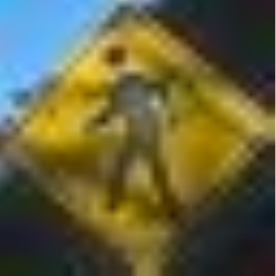} & \includegraphics[height=0.8cm]{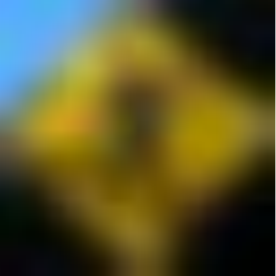} \\
FGSM \cite{goodfellow2014explaining}             & 13.26           & 31.80\%         &\includegraphics[height=0.8cm]{fig-pedestrian-clean.pdf}   & --- & \includegraphics[height=0.8cm]{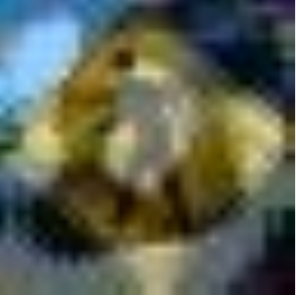} & \includegraphics[height=0.8cm]{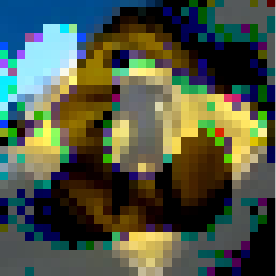} \\
Pointwise \cite{schott2018towards}              & 2.47               & 20.30\% &\includegraphics[height=0.8cm]{fig-pedestrian-clean.pdf}   & \includegraphics[height=0.8cm]{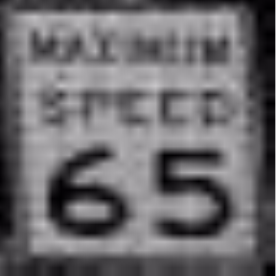} & \includegraphics[height=0.8cm]{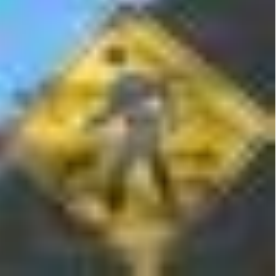} & \includegraphics[height=0.8cm]{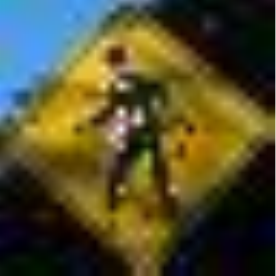} \\
Boundary attack \cite{brendel2018decisionbased}  & 2.42   & 22.10\%         &\includegraphics[height=0.8cm]{fig-pedestrian-clean.pdf}   & \includegraphics[height=0.8cm]{fig-SpeedLimit65.pdf} & \includegraphics[height=0.8cm]{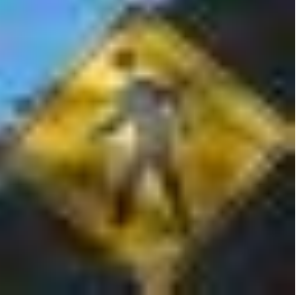} & \includegraphics[height=0.8cm]{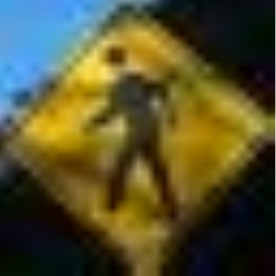} \\
$\text{RP}_\text{2}$ \cite{eykholt2018robust}                     & 10.73           & 88.90\%         &\includegraphics[height=0.8cm]{fig-pedestrian-clean.pdf}   & \includegraphics[height=0.8cm]{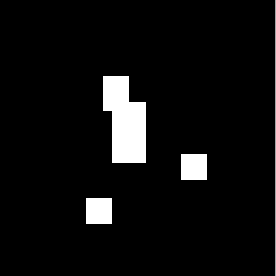} & \includegraphics[height=0.8cm]{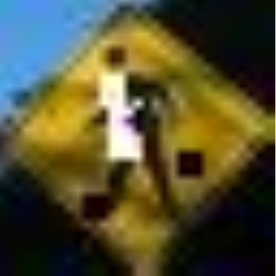}  & --- \\
Our TAA                                          & 7.47            & \textbf{99.10\%}  &\includegraphics[height=0.8cm]{fig-pedestrian-clean.pdf}   & \includegraphics[height=0.8cm]{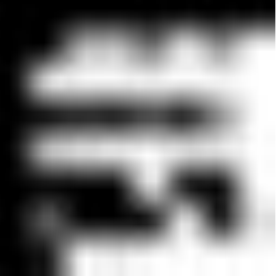} & \includegraphics[height=0.8cm]{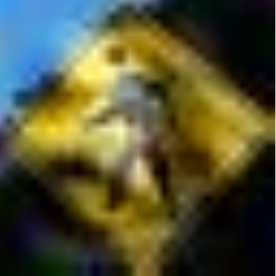} & --- \\
\bottomrule
\end{tabular}
\end{center}
\end{table*}

\begin{table*}[t]
\begin{center}
\setlength{\abovecaptionskip}{2pt}
\caption{Transfer attack on GTSRB dataset.}
\label{tab-result-GTSRB}
\begin{tabular}{m{1.5cm}<{\centering}m{1.1cm}<{\centering}m{1.1cm}<{\centering}m{1.2cm}<{\centering}m{1.2cm}<{\centering}m{1.1cm}<{\centering}m{1.1cm}<{\centering}m{1.2cm}<{\centering}m{1.2cm}<{\centering}}
\toprule\noalign{\smallskip}
\multirow{2}[3]{*}{Methods} & \multicolumn{4}{c}{Stop-SpeedLimit45} & \multicolumn{4}{c}{PedestrainCrossing-SpeedLimit65} \\
\cmidrule(lr){2-5} \cmidrule(lr){6-9}
~                           & $P_{loss}$ & $ASR$ & Original & Adv. & $P_{loss}$ & $ASR$ & Original & Adv. \\
\noalign{\smallskip}
\midrule
\noalign{\smallskip}
$\text{RP}_\text{2}$ \cite{eykholt2018robust} & 10.81 & 53.80\%  &\includegraphics[height=0.8cm]{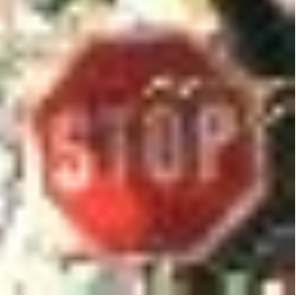}   &  \includegraphics[height=0.8cm]{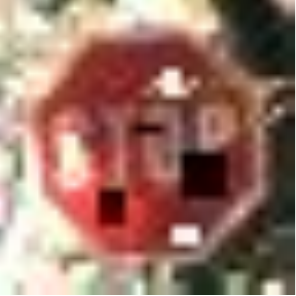} & 10.73 & 18.30\%  &\includegraphics[height=0.8cm]{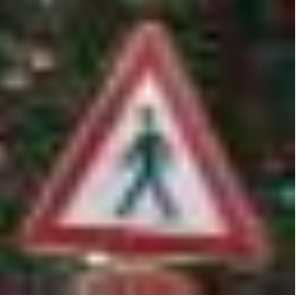}   & \includegraphics[height=0.8cm]{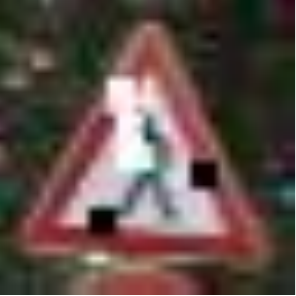} \\
Our TAA         & \textbf{7.62}  & \textbf{86.70\%}  &\includegraphics[height=0.8cm]{fig-GTSRB-Clean-Stop.pdf}   & \includegraphics[height=0.8cm]{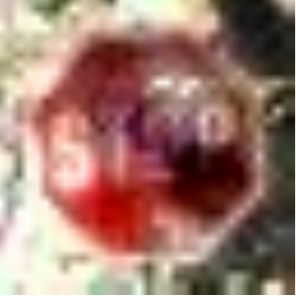} & \textbf{7.47}  & \textbf{36.70\%}  &\includegraphics[height=0.8cm]{fig-GTSRB-Pedestrian-clean.pdf}   & \includegraphics[height=0.8cm]{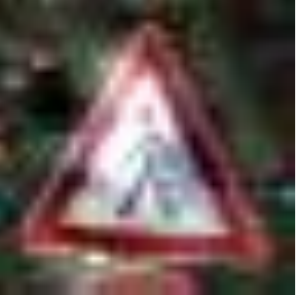}\\
\bottomrule
\end{tabular}
\end{center}
\end{table*}

\subsubsection{Robustness and Efficiency}
\begin{figure}[t]
\centerline{\includegraphics[height=7.3cm]{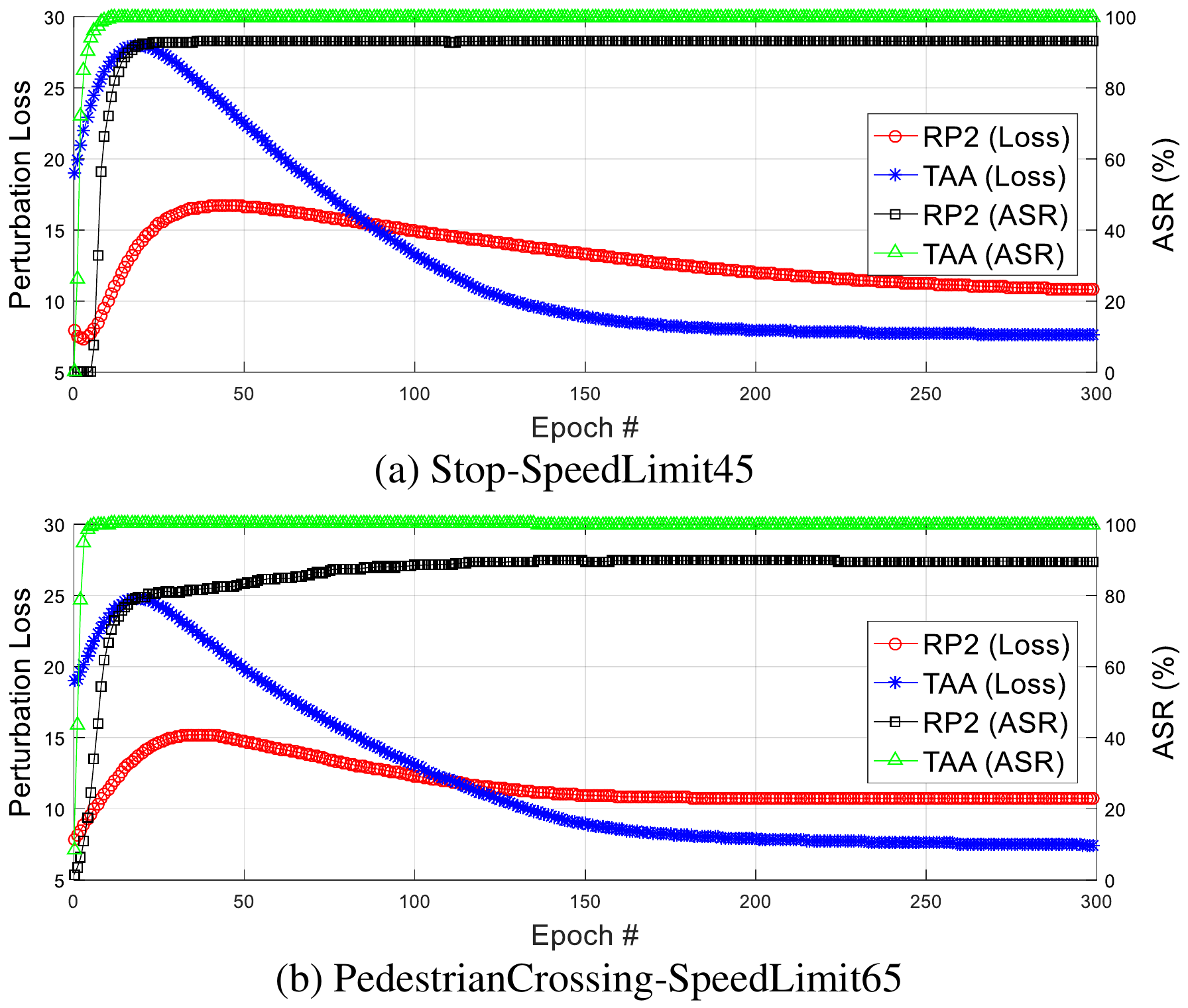}}
\caption{Robustness test under a wide range of training epochs.}
\label{fig-Epoch-Loss-ASR}
\end{figure}

We analysis the robustness of our TAA on the training set during the perturbations optimization steps. Fig.~\ref{fig-Epoch-Loss-ASR} illustrates the attacking performance of our TAA and $\text{RP}_\text{2}$ under different training epochs. It can be seen from Fig.~\ref{fig-Epoch-Loss-ASR} that our TAA continues to gain significant improvements in a large scale with respect to both $ASR$ and $P_{loss}$. For the $ASR$, our TAA converges much faster than $\text{RP}_\text{2}$ and performs more stable after convergence. For the $P_{loss}$, TAA shows a greater degree of flexibility before $150$ epochs and outperforms $\text{RP}_\text{2}$ by a large margin when epoch $>150$.

\begin{figure*}[t]
\captionsetup{justification=centering}
\centerline{\includegraphics[height=1.6cm]{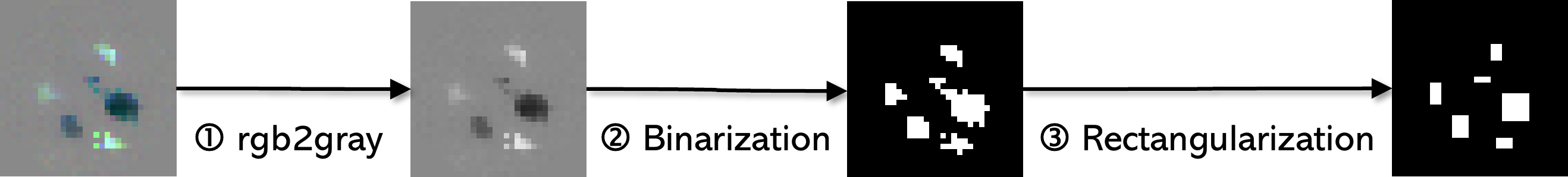}}
\caption{The $\text{RP}_\text{2}$ mask is crafted by several manually intervened post-processing steps.}
\label{fig-processL1mask}
\end{figure*}

It is worth mentioning that the $\text{RP}_\text{2}$ mask learned by $L_1$-norm originally appears irregular shapes (see the left most picture in Fig.~\ref{fig-processL1mask}). Thereby, it has to go through a few manual steps, e.g., binarization and rectangularization, which are easily influenced by people's subjective consciousness. Additionally, it should be relearned if the source or target class has been changed, while all of our soft attention maps could be built at once after the RAN training. The fully automatic learning procedure and the one-shot learning strategy are two main factors that improving the efficiency of our TAA.

\subsubsection{Transferability}
We test the data transferability and model transferability of TAA, separately. For data transfer, we apply the perturbation learned from LISA dataset to attack the traffic signs of GTSRB, and check whether the classifier will be misled. The attack results are listed in Table~\ref{tab-result-GTSRB}. For ``Stop-SpeedLimit45" attack, we can see that our TAA achieves much higher attack success accuracy ($86.7\%$) than $\text{RP}_\text{2}$ ($53.8\%$), i.e., more than $30\%$. For the ``PedestrianCrossing-SpeedLimit65" attack, it can be found that both $\text{RP}_\text{2}$ and TAA have a weak attack performance. The reason is that the ``PedestrianCrossing" sign of US (LISA) and German (GTSRB) are greatly different in both shape and color, i.e., only the middle pedestrian area looks similar. Although this makes the transfer attack harder, our TAA gets almost double $ASR$ compared with $\text{RP}_\text{2}$. This means the TAA perturbation has a much better transferability than $\text{RP}_\text{2}$ in attacking unknown data, even if the data lies in diverse distribution space.

To investigate the model transferability, we train three proximity DNNs classifiers, referred as CNN-2, CNN-3 and CNN-4. Compared to the original CNN, CNN-2 has an additional convolutional layer and nonlinear processing layer, CNN-3 adds one more nonlinear processing layer, and CNN-4 replaces all the ``Relu" activation function with ``tanh". Table~\ref{tab-result-ModelTransfer} lists their classification performance on clean input as well as the transfer attack success rate. We can see from Table~\ref{tab-result-ModelTransfer} that TAA achieves higher $ASR$ than $\text{RP}_\text{2}$ in most cases. This indicates our TAA method is more powerful for real hackers, who do not know the target model, and thus, usually depend on the model transfer attack.
\begin{table*}[t]
\begin{center}
\caption{Transfer attack on various of DNNs models.}
\label{tab-result-ModelTransfer}
\begin{tabular}{m{1.4cm}<{\centering}m{1.5cm}<{\centering}m{1.5cm}<{\centering}m{1.55cm}<{\centering}m{1.55cm}<{\centering}m{1.6cm}<{\centering}m{1.6cm}<{\centering}}
\toprule\noalign{\smallskip}
\multirow{2}[3]{*}{Models} & \multirow{2}[3]{*}{Train Acc.} & \multirow{2}[3]{*}{Test Acc.} & \multicolumn{2}{c}{Stop-SpeedLimit45} & \multicolumn{2}{c}{Pedestrain-SpeedLimit65} \\
\cmidrule(lr){4-5} \cmidrule(lr){6-7}
~        &~         &~        & $\text{RP}_\text{2}$ \cite{eykholt2018robust} & Ours TAA & $\text{RP}_\text{2}$ \cite{eykholt2018robust} & Our TAA  \\
\noalign{\smallskip}
\midrule
\noalign{\smallskip}
CNN-2    & 91.20\% & 89.10\%  & \textbf{77.20\%}   & 65.10\%    & 38.70\%   & \textbf{59.90\%}  \\
CNN-3    & 94.60\% & 93.90\%  & 73.90\%   & \textbf{89.30\%}    & 12.40\%   & \textbf{41.00\%}  \\
CNN-4    & 96.20\% & 95.10\%  & 88.70\%   & \textbf{95.90\%}    & 38.20\%   & \textbf{74.20\%}  \\
\bottomrule
\end{tabular}
\end{center}
\end{table*}

\begin{table*}[t]
\begin{center}
\caption{Generalization test on more attack scenarios.}
\label{table:generalization}
\begin{tabular}{L{4cm}m{1.2cm}<{\centering}m{1.2cm}<{\centering}m{1.2cm}<{\centering}m{1.2cm}<{\centering}m{1.2cm}<{\centering}m{1.2cm}<{\centering}}
\toprule\noalign{\smallskip}
Input$\rightarrow$Target & $P_{loss}$ & $ASR$ & Original & Mask & Adv. & Noises \\
\noalign{\smallskip}
\midrule
\noalign{\smallskip}
Stop$\rightarrow$TurnRight                & 8.42      & 99.70\% &\includegraphics[height=0.8cm]{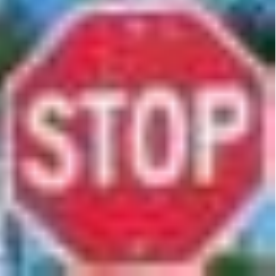} & \includegraphics[height=0.8cm]{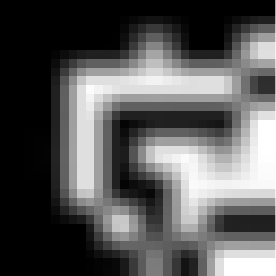} & \includegraphics[height=0.8cm]{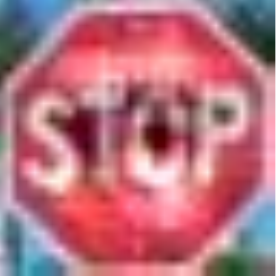} & \includegraphics[height=0.8cm]{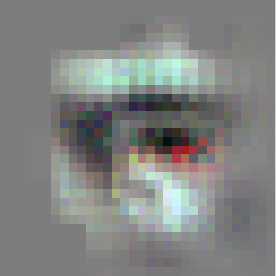}\\
PedestrianCrossing$\rightarrow$Merge      & 5.02      & 99.10\% &\includegraphics[height=0.8cm]{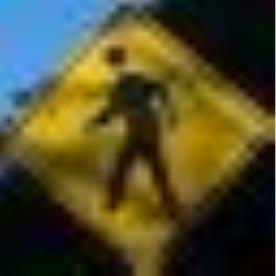} & \includegraphics[height=0.8cm]{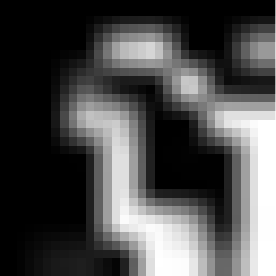} & \includegraphics[height=0.8cm]{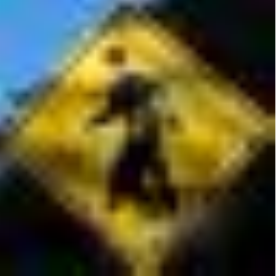} & \includegraphics[height=0.8cm]{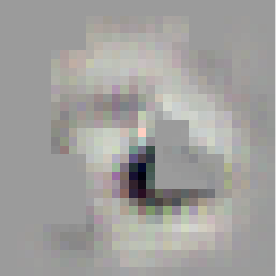}\\
SignalAhead$\rightarrow$TurnRight         & 6.61      & 100\%   &\includegraphics[height=0.8cm]{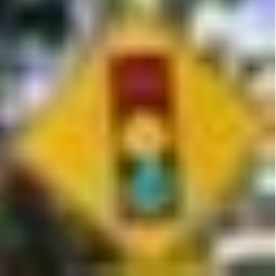}   & \includegraphics[height=0.8cm]{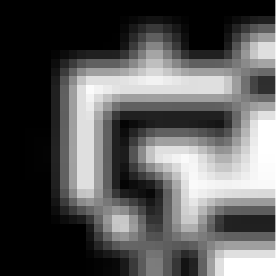} & \includegraphics[height=0.8cm]{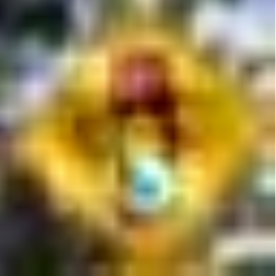} & \includegraphics[height=0.8cm]{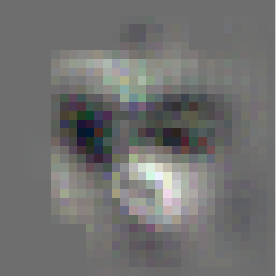} \\
SpeedLimit35$\rightarrow$School           & 4.43      & 100\%   &\includegraphics[height=0.8cm]{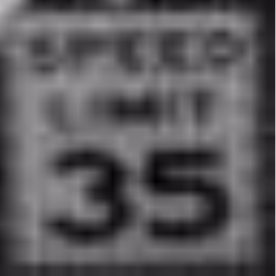}   & \includegraphics[height=0.8cm]{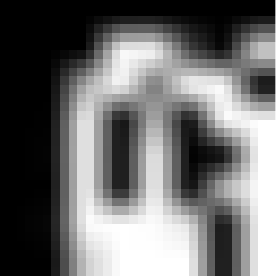} & \includegraphics[height=0.8cm]{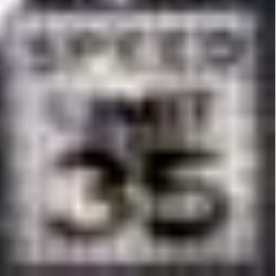} & \includegraphics[height=0.8cm]{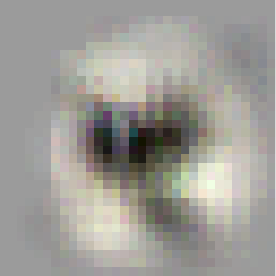} \\
SpeedLimit25$\rightarrow$NoLeftTurn       & 4.36      & 100\%   &\includegraphics[height=0.8cm]{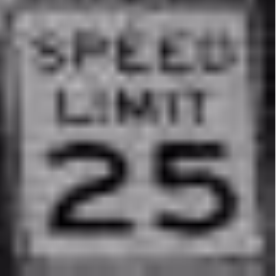}   & \includegraphics[height=0.8cm]{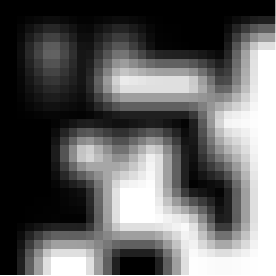} & \includegraphics[height=0.8cm]{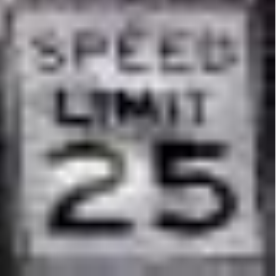} & \includegraphics[height=0.8cm]{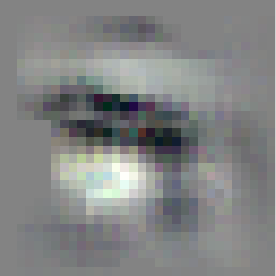} \\
KeepRight$\rightarrow$Yield               & 5.07      & 100\% &\includegraphics[height=0.8cm]{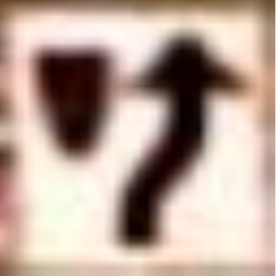} & \includegraphics[height=0.8cm]{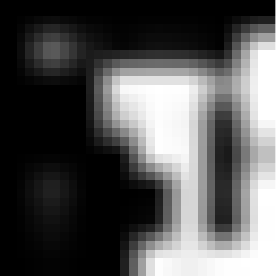} & \includegraphics[height=0.8cm]{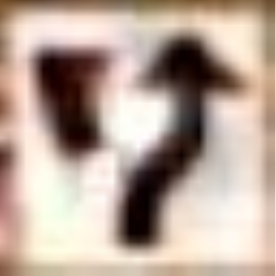} & \includegraphics[height=0.8cm]{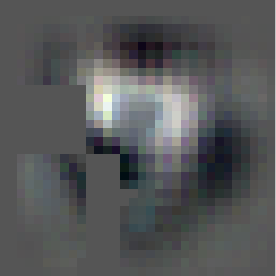}\\
AddedLine$\rightarrow$SpeedLimit40        & 8.11      & 100\% &\includegraphics[height=0.8cm]{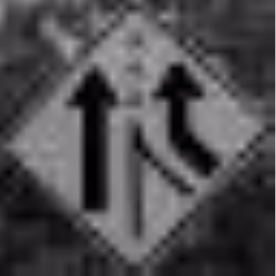} & \includegraphics[height=0.8cm]{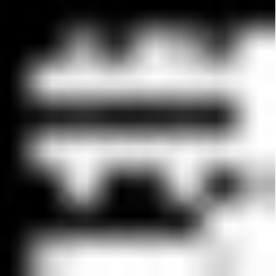} & \includegraphics[height=0.8cm]{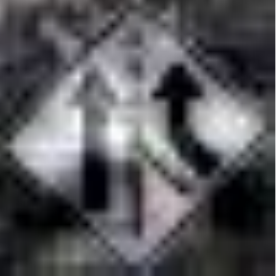} & \includegraphics[height=0.8cm]{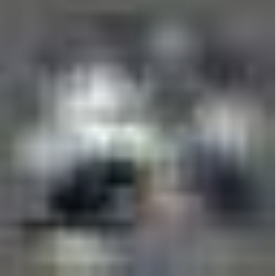}\\
Merge$\rightarrow$NoLeftTurn              & 5.61      & 100\% &\includegraphics[height=0.8cm]{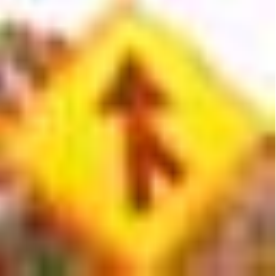} & \includegraphics[height=0.8cm]{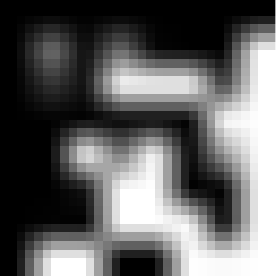} & \includegraphics[height=0.8cm]{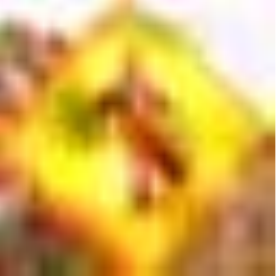} & \includegraphics[height=0.8cm]{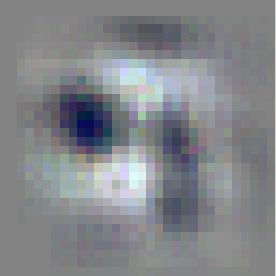}\\
Yield$\rightarrow$RoundAbout              & 6.10      & 100\% &\includegraphics[height=0.8cm]{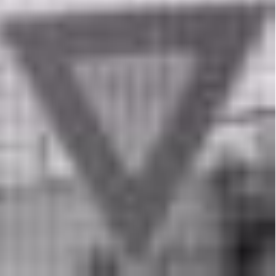} & \includegraphics[height=0.8cm]{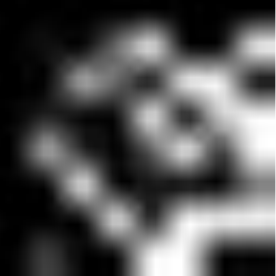} & \includegraphics[height=0.8cm]{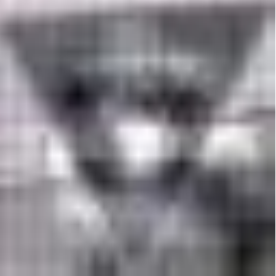} & \includegraphics[height=0.8cm]{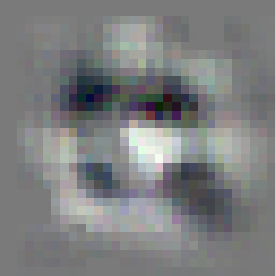}\\
LaneEnds$\rightarrow$RightLaneMustTurn    & 4.73      & 100\% &\includegraphics[height=0.8cm]{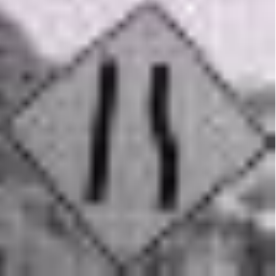} & \includegraphics[height=0.8cm]{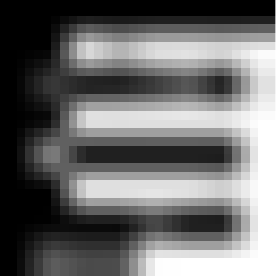} & \includegraphics[height=0.8cm]{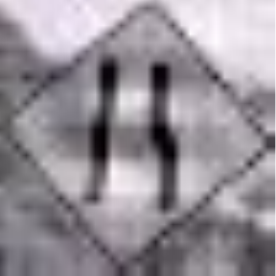} & \includegraphics[height=0.8cm]{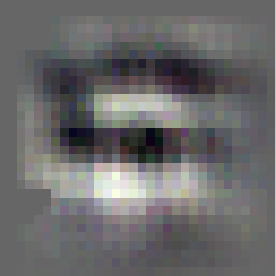}\\
StopAhead$\rightarrow$SchoolSpeedLimit25  & 6.39      & 100\% &\includegraphics[height=0.8cm]{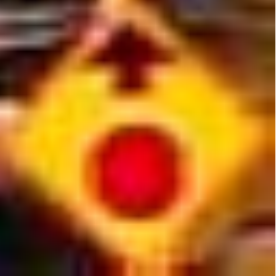} & \includegraphics[height=0.8cm]{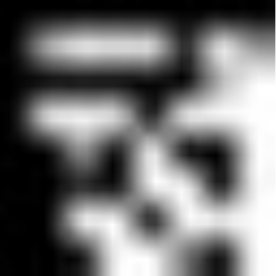} & \includegraphics[height=0.8cm]{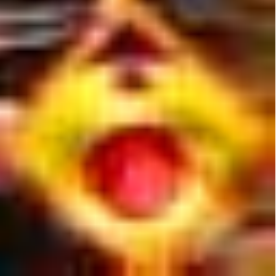} & \includegraphics[height=0.8cm]{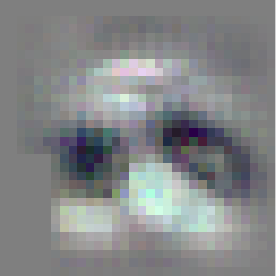}\\
\bottomrule
\end{tabular}
\end{center}
\end{table*}

\subsubsection{Generalization Capability} We validate the generalization capability of TAA on eleven more kind of road sign attacks (as shown in Table~\ref{table:generalization}). For each input class, $80\%$ images are used for training and $20\%$ for testing. Table~\ref{table:generalization} elaborates that TAA accomplishes nearly perfect $ASR$ by adding breezing and soft noises. These noises are even close to those generated by a one-to-one attack algorithms in terms of the perturbation loss, but fool the CNN model in a high rate. The strong generalization ability makes our TAA more valuable for malicious attackers. On the other hand, we hope TAA can inspire more efforts in designing robust defense systems to address such potential problems.

\subsubsection{Physical World Attack}
Finally, we test the robustness of $\text{RP}_\text{2}$ and TAA on physical world attack scenarios by printing their perturbations on cardboards. Notably, both $\text{RP}_\text{2}$ and TAA noises are first resized via bilinear interpolation to match the real signpost. After enlargement, $\text{RP}_\text{2}$ loss and TAA loss become $388.78$ and $292.91$, respectively. Then we place the attacked signs in four different scenes, i.e., three outdoors to imitate different sunlight conditions and one indoor scene to simulate the indoor parking area. For each scene, we take at least $10$ photos from different distances (i.e., from 1m to 10m for outdoor tests, and from 1m to 15m for indoor test) with a randomly selected view angle, as shown in Table \ref{tab-Physical}. Considering the distance, view angle and sunlight are most changed factors in physical world road condition, we believe these four scenarios are sufficient to cover the requirements of the practical application.

\begin{table*}[t]
\begin{center}
\caption{Physical world attack. Our TAA achieves high $ASR$ with much less $P_{loss}$.}
\label{tab-Physical}
\begin{tabular}{m{1.5cm}<{\centering}|m{1.5cm}<{\centering}m{1.5cm}<{\centering}|m{1.5cm}<{\centering}m{1.5cm}<{\centering}| m{1.5cm}<{\centering}m{1.5cm}<{\centering}|m{1.5cm}<{\centering}m{1.5cm}<{\centering}}
\toprule\noalign{\smallskip}
Scenes  & \multicolumn{2}{c|}{Park ($20^\circ$)} & \multicolumn{2}{c|}{Sunshine ($0^\circ$)} & \multicolumn{2}{c|}{Cloudy ($29^\circ$)} & \multicolumn{2}{c}{Indoor ($0^\circ$)} \\
\noalign{\smallskip}
\hline
\noalign{\smallskip}
1m      & \includegraphics[height=1.8cm]{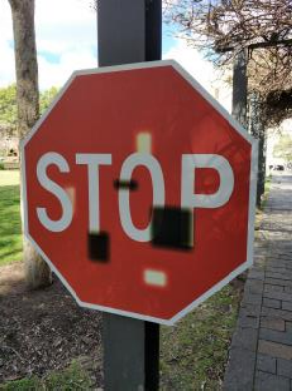}   &  \includegraphics[height=1.8cm]{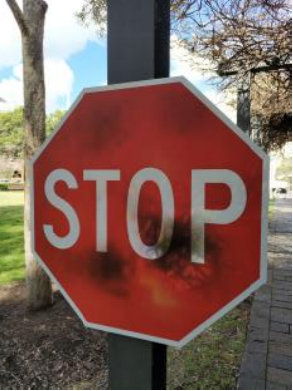} &
\includegraphics[height=1.8cm]{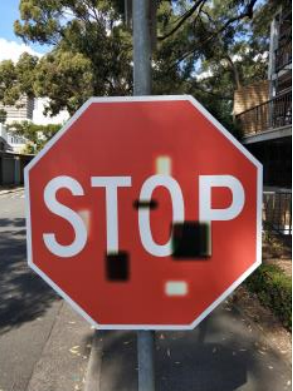} &
\includegraphics[height=1.8cm]{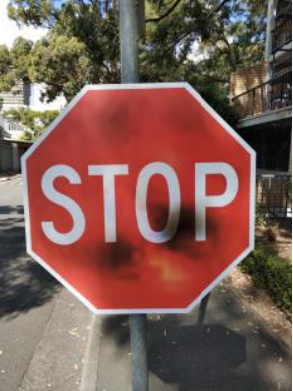} &
\includegraphics[height=1.8cm]{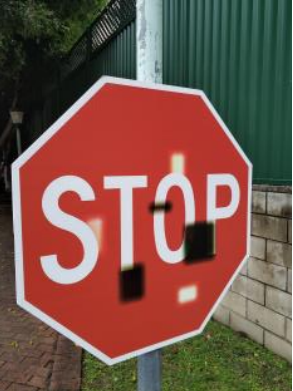} &
\includegraphics[height=1.8cm]{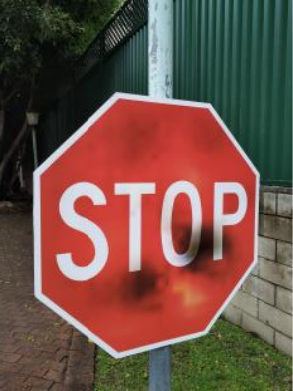} &
\includegraphics[height=1.8cm]{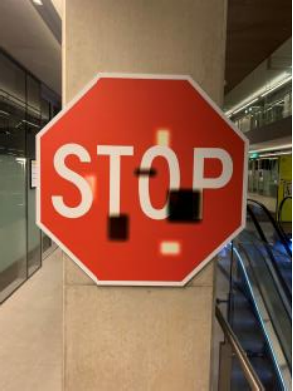} &
\includegraphics[height=1.8cm]{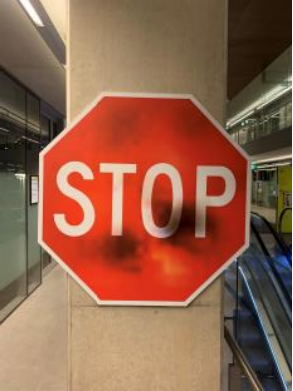}\\
5m      & \includegraphics[height=1.8cm]{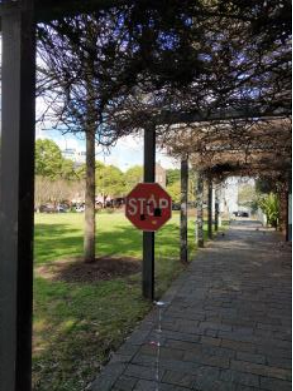}   & \includegraphics[height=1.8cm]{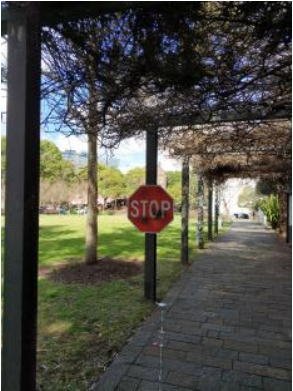} &
\includegraphics[height=1.8cm]{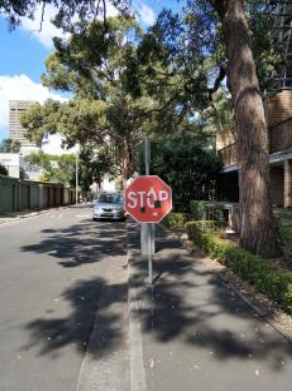} &
\includegraphics[height=1.8cm]{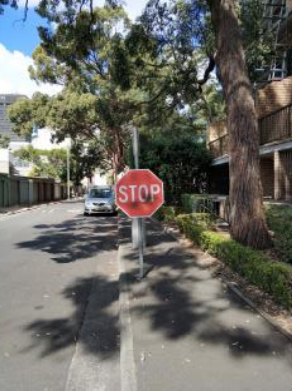} &
\includegraphics[height=1.8cm]{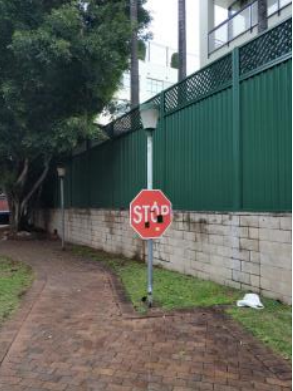} &
\includegraphics[height=1.8cm]{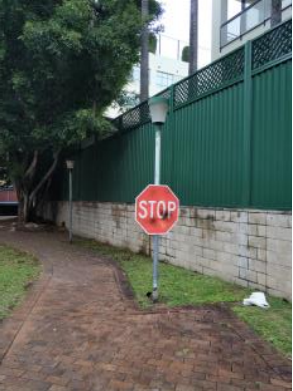} &
\includegraphics[height=1.8cm]{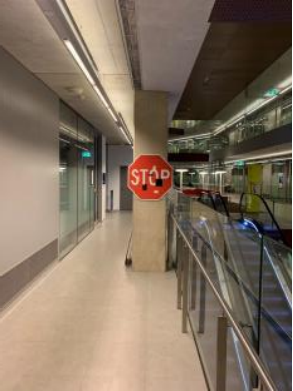} &
\includegraphics[height=1.8cm]{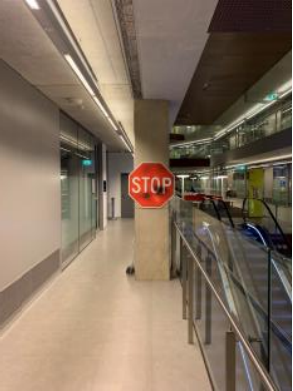}\\
10m     & \includegraphics[height=1.8cm]{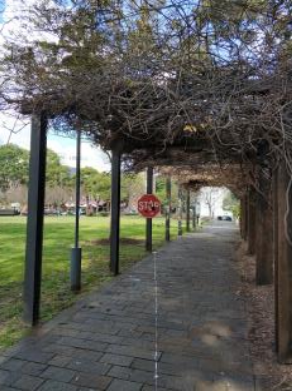}   & \includegraphics[height=1.8cm]{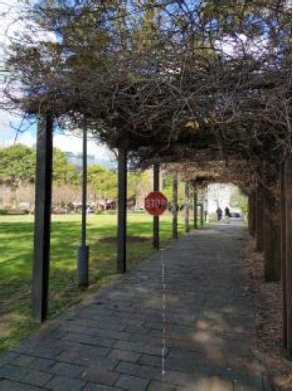} &
\includegraphics[height=1.8cm]{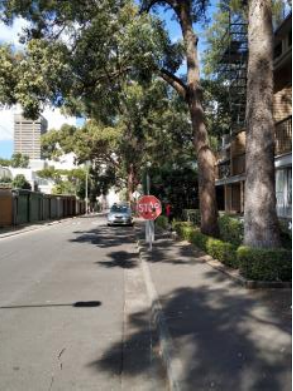} &
\includegraphics[height=1.8cm]{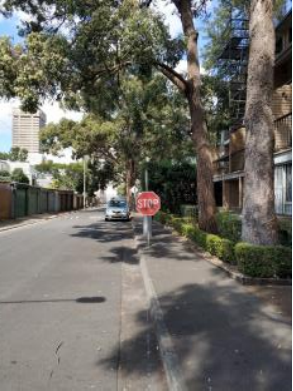} &
\includegraphics[height=1.8cm]{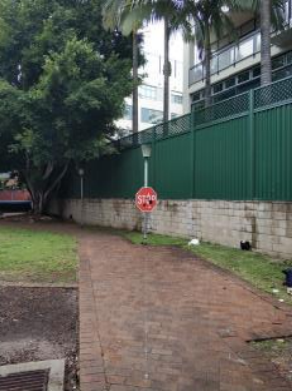} &
\includegraphics[height=1.8cm]{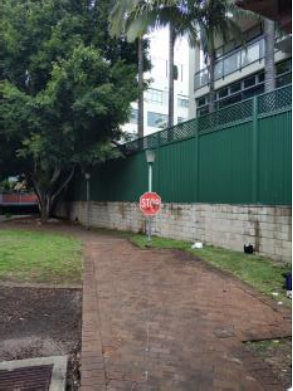} &
\includegraphics[height=1.8cm]{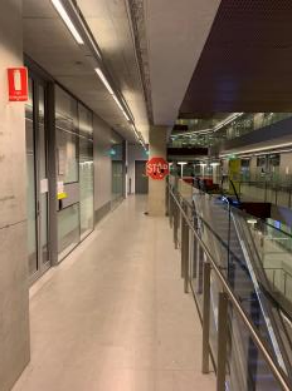} &
\includegraphics[height=1.8cm]{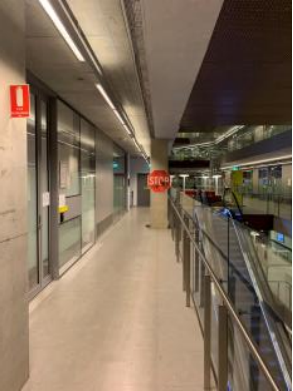}\\
\midrule
Methods &  $\text{RP}_\text{2}$     &TAA       &  $\text{RP}_\text{2}$       & TAA       &  $\text{RP}_\text{2}$       & TAA        &  $\text{RP}_\text{2}$       & TAA \\
$ASR$     & $90\%$   & $90\%$   & $100\%$    & $100\%$   & $80\%$     & $90\%$     & $100\%$    & $100\%$\\
$P_{loss}$    & $388.78$ & $292.91$ & $388.78$   & $292.91$  & $388.78$   & $292.91$   & $388.78$   & $292.91$ \\
\bottomrule
\end{tabular}
\end{center}
\end{table*}

From Table \ref{tab-Physical}, we can get five conclusions. (1) The attack successful rate are very high ($100\%$) for both $\text{RP}_\text{2}$ and TAA when the road sign is photographed in $0^\circ$. (2) The performance of both algorithms decreased when the traffic sign are viewed from a certain angle, i.e., $20^\circ$ or $29^\circ$; (3) Our TAA achieves the comparable performance with $\text{RP}_\text{2}$ in three scenes (Park, Sunshine and Indoor) and outperforms $\text{RP}_\text{2}$ when the weather is cloudy. (4) Owing to the soft attention map, our TAA uses much less perturbations than PR2 (reduced nearly $25\%$) and yet looks more natural. (5) Since our attack can be disguised as shades and sunlights (e.g., similar to camouflages), it is harder to be found by maintenance workers.

\section{Conclusion}
\label{Conclusion}
Real world road sign recognition is an important step for autonomous vehicles, so it is necessary to conduct the security test via attack methods. In this paper, we proposed an targeted attention attack (TAA), which employs the residual attention network (RAN) to find the sensitive pixels and optimizes an universal perturbation. This perturbation is easily ignored by human driver but leads to a high fooling rate on a set of test images. Finally, we evaluate the effectiveness of TAA on the LISA dataset, GTSRB dataset and real-world road sign images. Experimental results show that TAA achieves a higher attack success rate and uses a less perturbation, compared with several baseline methods. In the future, research on improving the attention mechanism and defending these attention based attacks can be promising work directions.

% if have a single appendix:
%\appendix[Proof of the Zonklar Equations]
% or
%\appendix  % for no appendix heading
% do not use \section anymore after \appendix, only \section*
% is possibly needed

% use appendices with more than one appendix
% then use \section to start each appendix
% you must declare a \section before using any
% \subsection or using \label (\appendices by itself
% starts a section numbered zero.)
%

\begin{comment}
\appendices
\section{Proof of the First Zonklar Equation}
Appendix one text goes here.

% you can choose not to have a title for an appendix
% if you want by leaving the argument blank
\section{}
Appendix two text goes here.

% use section* for acknowledgment
\section*{Acknowledgment}

The authors would like to thank...
\end{comment}

% Can use something like this to put references on a page
% by themselves when using endfloat and the captionsoff option.
\ifCLASSOPTIONcaptionsoff
  \newpage
\fi

% trigger a \newpage just before the given reference
% number - used to balance the columns on the last page
% adjust value as needed - may need to be readjusted if
% the document is modified later
%\IEEEtriggeratref{8}
% The "triggered" command can be changed if desired:
%\IEEEtriggercmd{\enlargethispage{-5in}}

% references section

% can use a bibliography generated by BibTeX as a .bbl file
% BibTeX documentation can be easily obtained at:
% http://mirror.ctan.org/biblio/bibtex/contrib/doc/
% The IEEEtran BibTeX style support page is at:
% http://www.michaelshell.org/tex/ieeetran/bibtex/
%\bibliographystyle{IEEEtran}
% argument is your BibTeX string definitions and bibliography database(s)
%\bibliography{IEEEabrv,../bib/paper}
%
% <OR> manually copy in the resultant .bbl file
% set second argument of \begin to the number of references
% (used to reserve space for the reference number labels box)
\bibliographystyle{IEEEtran}
\bibliography{TAA-Reference}

% biography section
%
% If you have an EPS/PDF photo (graphicx package needed) extra braces are
% needed around the contents of the optional argument to biography to prevent
% the LaTeX parser from getting confused when it sees the complicated
% \includegraphics command within an optional argument. (You could create
% your own custom macro containing the \includegraphics command to make things
% simpler here.)
%\begin{IEEEbiography}[{\includegraphics[width=1in,height=1.25in,clip,keepaspectratio]{mshell}}]{Michael Shell}
% or if you just want to reserve a space for a photo:

\begin{IEEEbiography}[{\includegraphics[width=1in,height=1.25in,clip,keepaspectratio]{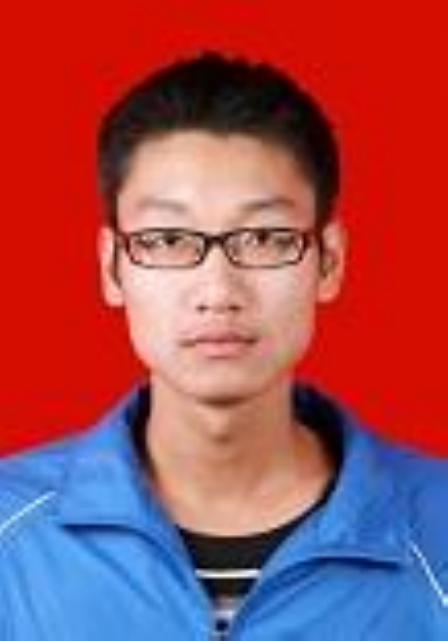}}]{Xinghao Yang}
received the B.Eng. degree in electronic information engineering and M.Eng. degree in information and communication engineering from the China University of Petroleum (East China), Qingdao, China, in 2015 and 2018, respectively. Currently, he is a PhD student in Advanced Analytics Institute, University of Technology Sydney, Australia.

His research interests include multi-view learning and adversarial machine learning with publications on information fusion and information sciences.
\end{IEEEbiography}
\vspace{-5ex}
\begin{IEEEbiography}[{\includegraphics[width=1in,height=1.25in,clip,keepaspectratio]{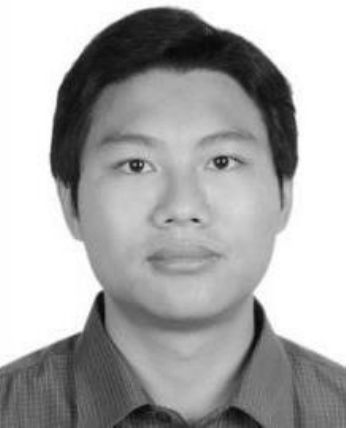}}]{Weifeng Liu}
(M'12-SM'17) received the double B.S. degrees in automation and business administration and the Ph.D. degree in pattern recognition and intelligent systems from the University of Science and Technology of China, Hefei, China, in 2002 and 2007, respectively.

He was a Visiting Scholar with the Centre for Quantum Computation and Intelligent Systems, Faculty of Engineering and Information Technology, University of Technology Sydney, Ultimo, NSW, Australia, from 2011 to 2012. He is currently a
Full Professor with the College of Information and Control Engineering, China University of Petroleum, Qingdao, China. He has authored or coauthored a dozen papers in top journals and prestigious conferences, including four Essential Science Indicators (ESI) highly cited papers and two ESI hot papers. His research interests include computer vision, pattern recognition, and machine learning.

Prof. Liu serves as an Associate Editor for the Neural Processing Letters, the Co-Chair for the IEEE SMC Technical Committee on Cognitive Computing, and a Guest Editor for special issue of the Signal Processing, theIET Computer
Vision, the Neurocomputing, and the Remote Sensing. He also serves over 20 journals and over 40 conferences.
\end{IEEEbiography}
\vspace{-5ex}
\begin{IEEEbiography}[{\includegraphics[width=1in,height=1.25in,clip,keepaspectratio]{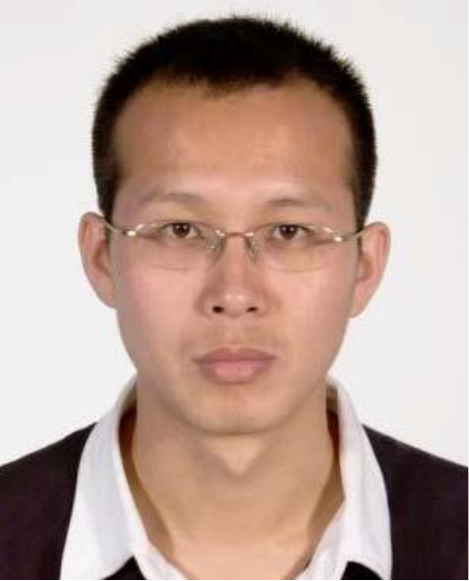}}]{Shengli Zhang}
(M'08-SM'16) received the B.Eng. degree in electronic engineering and the M.Eng. degree in communication and information engineering from the University of Science and Technology of China (USTC), Hefei, China, in 2002 and 2005, respectively, and the Ph.D. degree from the Department of Information Engineering, The Chinese University of Hong Kong (CUHK), in 2008.

After that, he joined the Communication Engineering Department, Shenzhen University. From 2013 to 2015, he was a Visiting Associate Professor with Stanford University. He is currently a Full Professor with the Communication Engineering Department, Shenzhen University. He is also the pioneer of Physical-layer network coding (PNC). He has published over 20 IEEE top journal papers and ACM top conference papers, including IEEE JSAC, IEEE TWC, IEEE TMC, IEEE TCom, and ACM Mobicom. His research interests include physical layer network coding, wireless networks, and blockchain.

Dr. Zhang has also severed as a TPC member in several IEEE conferences. He severed as an Editor for IEEE TVT, IEEE WCL, and IET Communications.
\end{IEEEbiography}
\vspace{-5ex}
\begin{IEEEbiography}[{\includegraphics[width=1in,height=1.25in,clip,keepaspectratio]{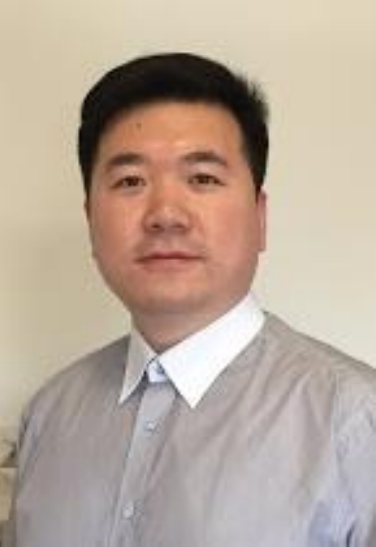}}]{Wei Liu}
is the Data Science Research Leader and a Senior Lecturer at the Advanced Analytics Institute, School of Software, Faculty of Engineering and Information Technology, the University of Technology, Sydney. Before joining UTS, he was a Research Fellow at the University of Melbourne and then a Machine Learning Researcher at NICTA. He obtained his PhD from the University of Sydney. He works in the areas of machine learning and data mining and has published over 80 papers in research topics of tensor factorization, adversarial learning, graph mining, causal inference, and anomaly detection. He has won 3 best paper awards.
\end{IEEEbiography}
\vspace{-5ex}
\begin{IEEEbiography}[{\includegraphics[width=1in,height=1.25in,clip,keepaspectratio]{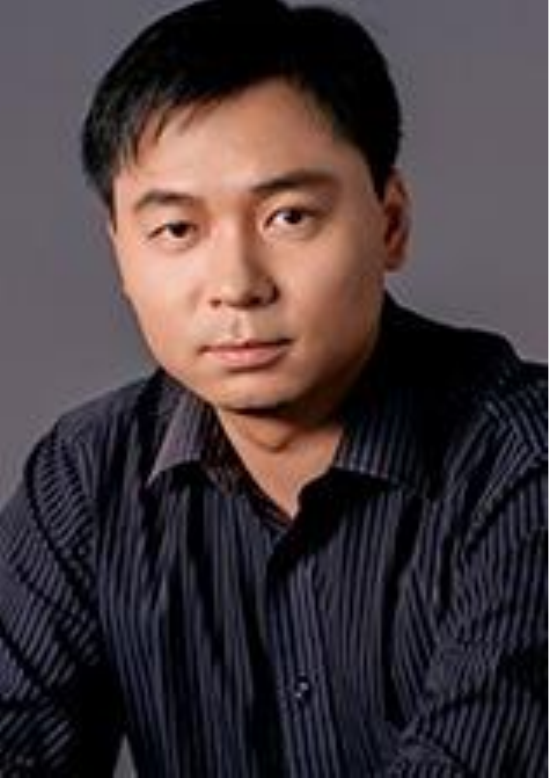}}]{Dacheng Tao}
(F'15) is Professor of Computer Science and ARC Laureate Fellow in the School of Computer Science and the Faculty of Engineering, and the Inaugural Director of the UBTECH Sydney Artificial Intelligence Centre, at The University of Sydney. He mainly applies statistics and mathematics to Artificial Intelligence and Data Science. His research results in artificial intelligence have expounded in one monograph and 200+ publications at prestigious journals and prominent conferences, such as IEEE T-PAMI, IJCV, JMLR, AAAI, IJCAI, NIPS, ICML, CVPR, ICCV, ECCV, ICDM, and KDD, with several best paper awards. He received the 2018 IEEE ICDM Research Contributions Award and the 2015 Australian Scopus-Eureka prize. He is a Fellow of the IEEE and the Australian Academy of Science.
\end{IEEEbiography}

% You can push biographies down or up by placing
% a \vfill before or after them. The appropriate
% use of \vfill depends on what kind of text is
% on the last page and whether or not the columns
% are being equalized.

%\vfill

% Can be used to pull up biographies so that the bottom of the last one
% is flush with the other column.
%\enlargethispage{-5in}

% that's all folks
\end{document}